\documentclass[conference]{IEEEtran}
\pdfoutput=1
\IEEEoverridecommandlockouts

\usepackage{cite}
\usepackage{amsmath,amssymb,amsfonts,bm}
\usepackage{algorithmic}
\usepackage[linesnumbered,ruled,noline,noend]{algorithm2e}
\usepackage{xcolor, soul}
\usepackage{graphicx}
\usepackage{textcomp}
\usepackage{color, colortbl}
\usepackage{spverbatim}
\usepackage{etoolbox}
\usepackage{multirow}
\usepackage{booktabs}
\usepackage{courier}
\usepackage{bm}
\usepackage{gensymb}
\usepackage{diagbox} 
\usepackage{bbm}
\usepackage{url}
\usepackage{array}
\usepackage[caption=false]{subfig}
\usepackage{xcolor}
\usepackage{caption,setspace} 
\usepackage{tabularx,booktabs}
\newcolumntype{b}{>{\centering\arraybackslash}X}
\newcolumntype{s}{>{\centering\arraybackslash\hsize=.67\hsize}X}
\newcolumntype{d}{>{\centering\arraybackslash\hsize=.52\hsize}X}
\newcolumntype{e}{>{\centering\arraybackslash\hsize=.58\hsize}X}
\newcolumntype{f}{>{\centering\arraybackslash\hsize=.29\hsize}X}
\newcolumntype{h}{>{\centering\arraybackslash\hsize=.4\hsize}X}
\newcolumntype{g}{>{\centering\arraybackslash\hsize=.8\hsize}X}

\newcommand\T{\rule{0pt}{2.6ex}}       % Top strut
\newcommand\B{\rule[-1.2ex]{0pt}{0pt}} % Bottom strut

\makeatletter\renewcommand\paragraph{\@startsection{paragraph}{4}{\z@}
  {.2em \@plus1ex \@minus.2ex}{-.5em}{\normalfont\normalsize\bfseries}}\makeatother

\def\BibTeX{{\rm B\kern-.05em{\sc i\kern-.025em b}\kern-.08em
    T\kern-.1667em\lower.7ex\hbox{E}\kern-.125emX}}
\begin{document}
\title{
Label-Efficient Self-Supervised Federated Learning for Tackling Data Heterogeneity in Medical Imaging}
\author{Rui~Yan,
        Liangqiong~Qu,
        Qingyue~Wei,
        Shih-Cheng~Huang,
        Liyue~Shen,\\
        Daniel~Rubin,
        Lei~Xing, 
        Yuyin~Zhou
\thanks{This work was supported by the National Institutes of Health (NIH) under Grant R01CA256890, Grant R01CA227713, Grant U01CA242879, Grant R01HL155410, and Grant R01LM012966.}
\thanks{R. Yan is with the Institute for Computational and Mathematical Engineering, Stanford University, Stanford, CA 94305 USA
    (e-mail: ruiyan@stanford.edu).}
\thanks{L. Qu is with the Department of Statistics and Actuarial Science and the Institute of Data Science, The University of Hong Kong, Hong Kong, 999077 (e-mail: liangqqu@hku.hk).}
\thanks{D.L. Rubin is with the Department of Biomedical Data Science, Stanford University, Stanford, CA 94305 USA 
    (e-mail:rubin@stanford.edu).}
\thanks{Q. Wei and L. Shen are with the Department of Electrical Engineering, Stanford University, Stanford, CA 94305 USA 
    (e-mail: qywei@stanford.edu; liyues@stanford.edu).}
\thanks{SC. Huang is with the Department of Biomedical Informatics, Stanford University, Stanford, CA 94305 USA
    (e-mail: mschuang@stanford.edu).}
\thanks{L. Xing is with the Department of Radiation Oncology, Stanford University, Stanford, CA 94305 USA 
    (e-mail: lei@stanford.edu).}
\thanks{Y. Zhou is with the Department of Computer Science and Engineering at University of California, Santa Cruz, CA 95064 
    (e-mail: zhouyuyiner@gmail.com).}}
    
\maketitle

\begin{abstract}
The collection and curation of large-scale medical datasets from multiple institutions is essential for training accurate deep learning models, but privacy concerns often hinder data sharing.
Federated learning (FL) is a promising solution that enables privacy-preserving collaborative learning among different institutions, but it generally suffers from performance deterioration due to heterogeneous data distributions and a lack of quality labeled data.
In this paper, we present a robust and label-efficient self-supervised FL framework for medical image analysis.
Our method introduces a novel Transformer-based self-supervised pre-training paradigm that pre-trains models directly on decentralized target task datasets using masked image modeling, to facilitate more robust representation learning on heterogeneous data and effective knowledge transfer to downstream models.
Extensive empirical results on simulated and real-world medical imaging non-IID federated datasets show that masked image modeling with Transformers significantly improves the robustness of models against various degrees of data heterogeneity. 
Notably, under severe data heterogeneity, our method, without relying on any additional pre-training data, achieves an improvement of 5.06\%, 1.53\% and 4.58\% in test accuracy on retinal, dermatology and chest X-ray classification compared to the supervised baseline with ImageNet pre-training. 
In addition, we show that our federated self-supervised pre-training methods yield models that generalize better to out-of-distribution data and perform more effectively when fine-tuning with limited labeled data, compared to existing FL algorithms. 
The code is available at \url{https://github.com/rui-yan/SSL-FL}.
\end{abstract}

\begin{IEEEkeywords}
Federated Learning, Self-supervised Learning, Vision Transformers, Data Efficiency
\end{IEEEkeywords}

\section{Introduction}
\label{sec:introduction}
%----------fig1 start --------
\begin{figure}[t]
\begin{center}
    \includegraphics[width=\linewidth]{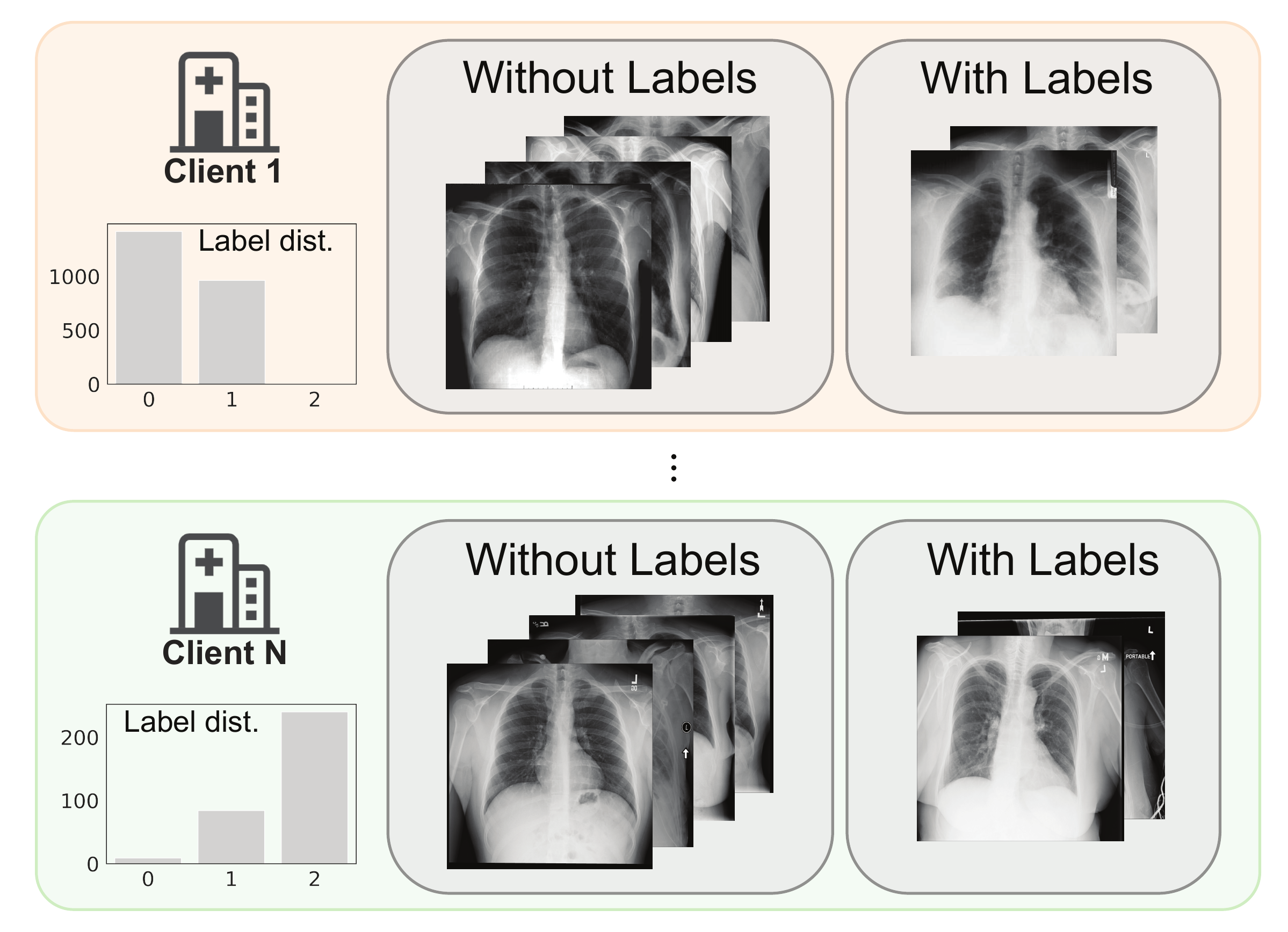}
\end{center}
\vspace{-1.2em}
\caption{Data heterogeneity and label deficiency of medical image datasets from different institutions.}
\vspace{-1.5em}
\label{Fig:Dataset}
\end{figure}
%----------fig1 end --------

\IEEEPARstart{F}{ederated} learning (FL) is a paradigm that allows model training using data distributed across multiple sites without explicit data sharing~\cite{mcmahan2017communication}. 
Compared to models trained at individual sites, federated models can be trained with a much more diverse and larger-scale dataset, which can result in superior performance and stronger generalizability. Therefore, this training paradigm has been widely adopted for critical medical applications such as the detection of brain tumors~\cite{sheller2020federated} and COVID-19~\cite{dayan2021federated,zhang2021dynamic}, and applied to various types of data, including medical imaging data, electronic health records and sensor data~\cite{sheller2020federated,kaissis2021end,rieke2020future}.

As a decentralized approach, FL suffers from performance degradation due to data heterogeneity and label deficiency~\cite{li2020federated2,rieke2020future, he2020group}. As shown in Fig.~\ref{Fig:Dataset}, data heterogeneity and label deficiency are particularly pronounced in medical image datasets of real-world applications. 
Regarding data heterogeneity, for example, some hospitals may have more data
from patients at an early stage while the others may collect the data with severe conditions only (\emph{i.e.}, label distribution skew). This is also referred to as statistical heterogeneity or non-identically distributed (non-IID) data partitions; 
large hospitals usually have more patient data than community clinics (\emph{i.e.}, quantity skew); 
and images at each hospital are acquired with different imaging acquisition protocols and on different patient populations (\emph{i.e.}, feature distribution skew). 
In terms of label deficiency, some sites may not have enough bandwidth or incentive for a complete labor-intensive data labeling and thus only a small proportion of all available medical images may be labeled.

While several research efforts~\cite{Li2020On, li2020federated, wang2020tackling, karimireddy2020scaffold} have been devoted to addressing the challenges caused by data heterogeneity, current approaches tend to deteriorate in performance when using strongly skewed data distributions~\cite{li2021model, zhao2018federated,qu2021rethinking}.
To handle extremely non-IID data partitions, recent studies~\cite{qu2021rethinking} suggest that Vision Transformers (ViTs)~\cite{dosovitskiy2021an} are better alternatives to convolution neural networks (CNNs), which have become the standard architecture used in the FL framework for image data.
Qu~\emph{et al.}~\cite{qu2021rethinking} reveal that
simply replacing CNNs with ViTs outperforms even the state-of-the-art optimization-based FL methods.
However, the success of such models largely relies on supervised ImageNet pre-training, which could suffer from domain discrepancy when fine-tuning with medical images and can be further improved by self-supervised pre-training on a centrally shared large-scale in-domain medical dataset~\cite{azizi2021big}. However, such centrally shared datasets rarely exist in the medical domain due to privacy and ownership concerns. Therefore, it is desired to build a self-supervised FL framework that collaboratively learns a global model by leveraging all available unlabeled data without sharing data among institutions.

Label deficiency is a common challenge in medical imaging. 
This can make it difficult to train accurate deep learning models on medical imaging data, as these models typically require large amounts of labeled data to learn effectively.
To address this issue, various approaches such as semi-supervised and self-supervised learning methods~\cite{zhang2021weakly,zhou2018brief,zhang2020weakly,zhou2022self,li2021rotation,azizi2021big} have been proposed to allow models to learn from partially labeled or unlabeled data.
However, many of these methods assume that the data is centralized, which are not practical for decentralized data.
To enable privacy-enhancing model training on decentralized medical data, Yang \emph{et al.}~\cite{yang2021federated} combine semi-supervised learning strategies such as consistency loss~\cite{berthelot2019mixmatch} with FL, referred to as Semi-FL.
Recently, several federated contrastive learning (FCL) methods~\cite{zhang2020federated,zhuang2021collaborative,zhuang2021divergence} have been proposed.
However, FCL methods often yield sub-optimal results when the data is highly heterogeneous or limited at local clients.

In this paper, we propose a robust self-supervised FL framework to address these challenges as shown in Fig.~\ref{Fig:framework}. 
To the best of our knowledge, this is the first work that simultaneously tackles the issues of data heterogeneity and label deficiency for medical imaging in FL leveraging masked image modeling as the self-supervised task.
Self-supervised pre-training has been demonstrated to be an effective solution to alleviate the need for large-scale labeled pre-training datasets and potentially generalizes better across various tasks~\cite{hendrycks2019using}. 
Moreover, unlike supervised learning which relies heavily on the label information, self-supervised pre-training learns the intrinsic features of images in local clients without labels, embodying less label-specific inductive bias and thus, less susceptible to label distribution skewness.
The proposed method learns visual representations more effectively across non-IID clients, even when data are limited at some clients.
To this end, we design a distributed self-supervised learning paradigm to improve FL in medical imaging, which consists of two essential steps: 
(1) \emph{federated self-supervised pre-training}, which exploits knowledge from decentralized unlabeled data based on masked image modeling in a distributed setting;
(2) \emph{federated supervised fine-tuning}, which then transfers this knowledge to the target tasks by fine-tuning the federated models.

Specifically, we implement two masked image modeling methods, BEiT~\cite{bao2021beit} and MAE~\cite{he2021masked}, as the SSL module in our federated framework. 
We evaluate their performance in both centralized and federated settings across diverse medical imaging tasks and demonstrate that BEiT and MAE together with Transformers are robust to distribution shifts and facilitate effective representation learning with limited amounts of data.

\vspace{1em}
We conduct extensive experiments under different degrees of data heterogeneity and with different fractions of labeled data on diverse medical datasets including diabetic retinopathy images, dermatology images and chest X-rays to validate the broad effectiveness of our self-supervised FL framework.

Our main contributions are summarized as follows:
\begin{itemize}
    \item We design a privacy-preserving federated self-supervised pre-training framework that uses masked image modeling to learn visual representations from decentralized data. Our proposed framework can tackle data heterogeneity and label deficiency at once.
    \item Our results on diverse medical datasets demonstrate that the proposed method is more label-efficient and robust to non-IID data compared to ImageNet supervised baselines and existing FL algorithms.
    \item For evaluation of a real-world distribution, we construct a federated chest X-rays benchmark called COVID-FL by curating the data from 8 different medical sites for testing the model's robustness in a realistic federated setting.

\end{itemize}

%======================== RELATED WORK =============================%
\section{related work}
\label{sec:relatedwork}
\subsection{Federated Learning}
Federated learning (FL) is a distributed training technique that trains machine learning models on private data across decentralized clients.
As a standard FL algorithm, FedAvg~\cite{mcmahan2017communication} performs local model training via stochastic gradient descent updates at each client, followed by a model aggregation at the server. 
However, FedAvg generally suffers from performance degradation due to the weight divergence issue caused by the non-IID data partitions (\emph{i.e.}, data heterogeneity)~\cite{hsieh2020non,li2020federated}. 
Many efforts have been devoted to addressing this issue, including regularizing the local model learning in parameter space~\cite{li2020federated, karimireddy2020scaffold}, 
sharing a subset of data among clients~\cite{zhao2018federated}, 
guiding local training with knowledge distillation~\cite{zhu2021data, lin2020ensemble}, and introducing effective global model aggregation strategies~\cite{wang2020tackling,Wang2020Federated}.
Recent studies tackle data heterogeneity from the perspective of model initialization, suggesting that pre-training alleviates the drastic accuracy drop caused by data heterogeneity~\cite{chen2022pre,nguyen2022begin}. Moreover, \cite{liang2020think} shows that Vision Transformer~\cite{dosovitskiy2021an} pre-trained on ImageNet leads to significant performance gain on non-IID data, outperforming its CNNs counterpart and optimization-based FL methods.
However, these FL methods assume fully labeled samples are available, which is not always feasible in the medical domain.
An FL method that could handle both data heterogeneity and limited annotations is desired.

\subsection{Self-supervised Learning}
Self-supervised learning (SSL), a method that exploits unlabeled data by using the data itself to provide the supervision, has gained popularity because of its ability to learn effective image representations~\cite{misra2020self, oord2018representation} and avoid the cost of annotating large-scale datasets. 
The core of SSL lies in adopting pretext tasks as self-supervision and then using the learned representations for different downstream tasks.
Various pretext tasks have been proposed such as image inpainting~\cite{pathak2016context} and jigsaw puzzle~\cite{noroozi2016unsupervised}, which have also been shown to be beneficial to medical imaging tasks~\cite{zhou2021models}.
A more recent research strand focuses on contrastive learning, which learns visual representations by forcing the representations of positive pairs to be closer while far apart for negative pairs.
To obtain enough informative negative pairs, contrastive learning methods such as MoCo~\cite{he2020momentum} and SimCLR~\cite{chen2020simple} rely on a large memory bank or batch size.
BYOL~\cite{grill2020bootstrap} instead trains one network to predict representations of the same image under a different augmented view obtained from the other network.

With the recent advance in Vision Transformer (ViT)~\cite{dosovitskiy2021an}, multiple works such as BEiT~\cite{bao2021beit} and MAE~\cite{he2021masked} have been proposed to learn visual representations by signal reconstruction given corrupted images.
We refer to this type of methods as masked image modeling. 
As opposed to contrastive learning, masked image modeling does not heavily depend on a large sample size or certain compositions of data augmentation while achieving competitive performance~\cite{he2021masked}.

\subsection{Federated Self-supervised Learning}

Federated self-supervised learning is referred to as applying self-supervised pre-training to learn representations from unlabeled decentralized data. It has attracted increasing attention in FL community given its capability to facilitate model learning when labeled data are limited at local clients.

Previous works~\cite{zhang2020federated,zhuang2021collaborative,zhuang2021divergence} mainly consider contrastive learning as the self-supervised task, known as federated contrastive learning (FCL). Note that contrastive learning needs large and diverse data to generate enough informative negative samples to train a good model, while a client in FL generally lacks of data diversity, especially for non-IID cases.
Regarding this issue, FedCA~\cite{zhang2020federated} shares local data features among clients. FedEMA~\cite{zhuang2021divergence} updates local models by using the exponential moving average of the global model. Several works~\cite{wu2021federated,dong2021federated} have been proposed for medical imaging tasks by using MoCo as the self-supervised method. To address the limited local data diversity, \cite{wu2021federated} shares local negative samples, and \cite{dong2021federated} shares the metadata of local image representations among clients.
Most FCL works~\cite{zhang2020federated,zhuang2021collaborative,zhuang2021divergence,dong2021federated}, however, consider only label distribution skew in their heterogeneity experiments, where the number of images at each client is sufficient (\emph{e.g.}, greater than 10000) and each client contains the same number of images.
It remains elusive how these FCL methods perform under highly heterogeneous data partitions with limited local data sample size (\emph{e.g.}, less than 200 images for certain clients).

In our work, we develop the first federated self-supervised pre-training framework that employs masked image modeling as the self-supervised task.
We demonstrate that masked image modeling (\emph{i.e.}, MAE and BEiT) coupled with Transformers is robust to distribution shifts and could learn effectively when data are relatively limited. The proposed pre-training scheme significantly advances the capability of federated models over highly heterogeneous data partitions.

%============================= SECTION 3: METHOD =============================%
\section{Methodology}
%+++++++++++++++++++++++ 3-1: METHOD-PROBLEM STATEMENT
\subsection{Problem Statement}
Our work aims at building a robust model that collaboratively learns from decentralized clients without data sharing. 
Specifically, our goal is to improve the model performance in FL especially for non-IID client data and in limited label scenarios.
Suppose there are $N$ clients. Each client $k\in\{1,..,N\}$ has a local dataset $\mathcal{D}^k$. To learn a generalized global model over $\mathcal{D}=\bigcup_{k=1}^N{\{\mathcal{D}^k\}}$, the global objective function is defined as follows:
\begin{equation} \label{eq:global_objective}
    \arg \min_w\mathcal{L}(w)=\sum_{k=1}^N\frac{|\mathcal{D}^k|}{|\mathcal{D}|}\mathcal{L}_k(w),
\end{equation}
and the local objective function $\mathcal{L}_k(w)$ in client $k$ measuring the local empirical loss over data distribution $\mathcal{D}^k$ is defined as:
\begin{equation} \label{eq:local_objective}
    \mathcal{L}_k(w)=\mathbb{E}_{x\sim\mathcal{D}^k}[\ell_k(w;\bm{x})],
\end{equation}
where $\ell_k$ is the loss function used for client $k$, and $w$ denotes the global model parameters to be learned.

The focus of our work is to address the data heterogeneity issue in FL, given that the data across different clients are usually non-IID,~\emph{i.e.}, $\mathcal{D}^m$ and $\mathcal{D}^n$ ($m\neq n$) follow different distributions $P_m(\bm{x}, y)$ and $P_n(\bm{x}, y)$. 
Furthermore, considering that some local clients may not have sufficient labeled data due to the lack of resources, in this paper, we also investigate how FL performs under limited annotation,~\emph{i.e.}, local dataset $\mathcal{D}^k$ comprises labeled data $\mathcal{D}^k_l=\{(\bm{x},y)\}$ and unlabeled data $\mathcal{D}^k_u=\{\bm{x}\}$, where $|\mathcal{D}^k_l|$ is relatively small.

%----------fig2 start --------
\vspace{-.5em}
\begin{figure*}[t]
\begin{center}
    \includegraphics[width=\linewidth]{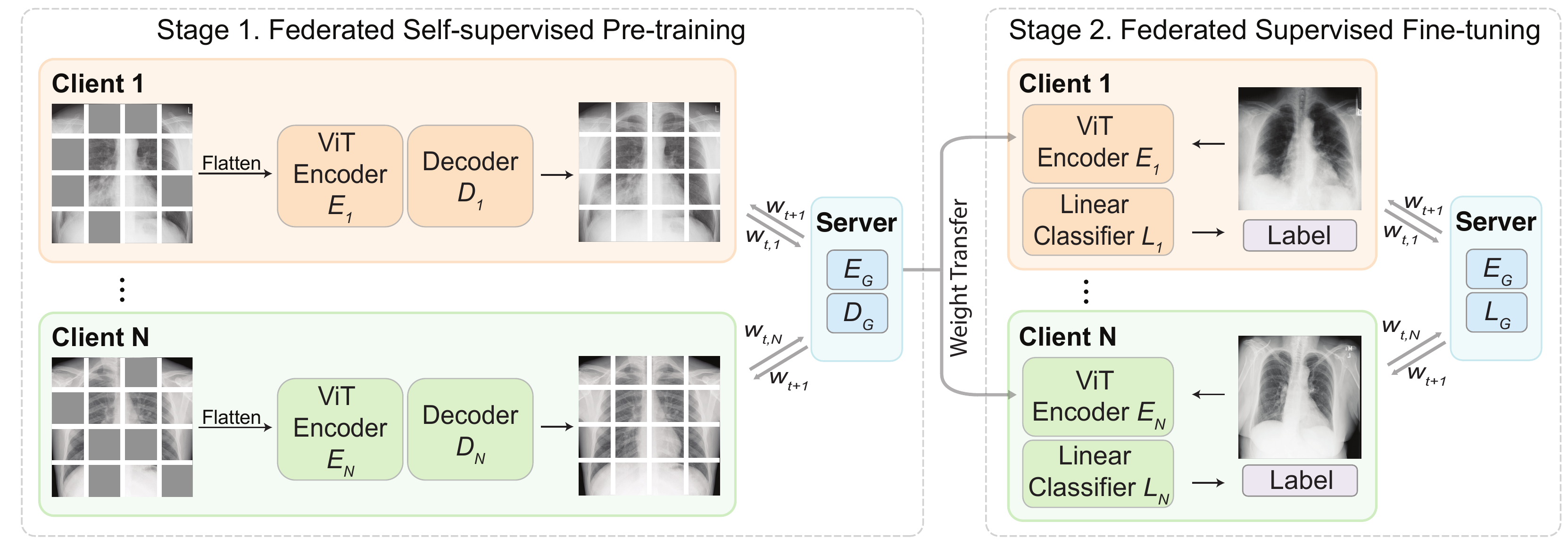}
\end{center}
\vspace{-1em}
\caption{Overview of the federated self-supervised learning framework.
In the pre-training stage (left), masked image modeling is used as the self-supervised task to learn representations from unlabeled images in each client. The pre-training process consists of three steps and ends when it reaches the maximum communication rounds $T$. 
At round $t$, (1) Each client $k$ ($k\in\{1,...,N\}$) trains its local auto-encoder $E_k$ and $D_k$ with the unlabeled local data; (2) Client $k$ uploads the weights of its auto-encoder $w_{t,k}$ to the central server; (3) The server produces a global auto-encoder $E_G$ and $D_G$ with weights $w_{t+1}$ via model weights averaging and broadcasts the global model back to each local client. 
In the fine-tuning stage (right), the final pre-trained global encoder $E_G^*$ from the first stage is used to initialize each local encoder $E_k$. A linear classifier $L_k$ is appended to each local encoder. End-to-end federated fine-tuning is performed on labeled images in each local client.} 
\label{Fig:framework}
% \vspace{-1em}
\end{figure*}
%----------fig2 end --------

%+++++++++++++++++++++++ 3-2: GENERALIZED FRAMEWORK
\subsection{Generalized Framework}
To address this important problem, we propose a generalized self-supervised FL framework to enhance both the robustness and the performance of federated models when learning from decentralized data with statistical heterogeneity. Our framework comprises two stages: a federated self-supervised pre-training stage and a supervised federated fine-tuning stage, as shown in Fig.~\ref{Fig:framework} and Alg.~\ref{algo:FL_pretraining}. During the self-supervised stage, the model exploits knowledge from decentralized data by pre-training with masked image modeling in a distributed setting. In the supervised federated fine-tuning stage, the knowledge is transferred from the previous stage to the target task by fine-tuning the federated models.

Specifically, we integrate two popular masked image modeling methods, BEiT\cite{bao2021beit} and MAE\cite{he2021masked}, into our generalized federated framework. We denote BEiT and MAE coupled with our framework as Fed-BEiT and Fed-MAE, respectively.
The pre-training and fine-tuning details of Fed-BEiT and Fed-MAE are illustrated in Sec.~\ref{sec:ssl pretraining} and Sec.~\ref{sec:supervised finetuning}.

%+++++++++++++++++++++++ 3-3: Self-supervised Federated Pre-training
\subsection{Federated Self-supervised Pre-training}
\label{sec:ssl pretraining}
During pre-training, the $k$-th local model is an autoencoder consisting of an encoder $E_k$ and a decoder $D_k$.
The model is trained using masked image modeling, which involves masking a subset of image patches and reconstructing the original signals in the masked patches. 
We implement two popular masked image modeling methods, BEiT~\cite{bao2021beit} and MAE~\cite{he2021masked}, as the SSL module in our federated framework. 
In this section, we describe the main components of our proposed federated pre-training protocol.

For the $k$-th client, an input image $\bm{x}\sim\mathcal{D}^k$ ($\bm{x}\in \mathbb{R}^{H\times W\times C}$) is divided into a sequence of image patches $\bm{x}_p=\{\bm{x}_p^i\}_{i=1}^P \in \mathbb{R}^{P\times (S^2 \cdot C)}$, where $(H,W)$ is the dimension of the original image, $C$ is the number of channels, $(S,S)$ is the dimension of each image patch and $P=HW/S^2$ is the number of image patches.

\subsubsection{Masking}
We denote the masking ratio as $\gamma$, the masked positions as $\mathcal{M}$, and the unmasked positions as $\mathcal{V}$. After randomly masking $\gamma\%$ of image patches, we get $|\mathcal{M}|=\gamma P$ and $|\mathcal{M}|+|\mathcal{V}|=P$. The total image patches can be represented as:
$
\bm{x}_p=\bm{x}_p^\mathcal{M}\cup\bm{x}_p^\mathcal{V}=\{\bm{x}_p^i, i\in \mathcal{M}\}\cup\{\bm{x}_p^i, i\in \mathcal{V}\},
$
where $\bm{x}_p^\mathcal{M}$ represents the masked patches and  $\bm{x}_p^\mathcal{V}$ represents the unmasked visible patches. Specifically, BEiT uses block-wise (n-gram) masking~\cite{bao2021beit}, and MAE uses random masking.

\subsubsection{Encoder}
We employ ViT~\cite{dosovitskiy2021an} as our encoder and apply it to a sequence of image patches as shown in Fig.~\ref{Fig:framework}.
\begin{itemize}
    \item For BEiT, the input to the ViT encoder is 
\[\{\bm{x}_p^\mathcal{V}\bm{E}\}\cup\{\bm{e}_p^\mathcal{M}\} \in\mathbb{R}^{P\times D}, \bm{E}\in\mathbb{R}^{(S^2\cdot C)\times D},
\]
 where $\bm{x}_p^\mathcal{V}\bm{E}\in\mathbb{R}^{|\mathcal{V}|\times D}$ is the linear projection of the visible patches to dimension $D$ and $\bm{e}_p^\mathcal{M}=\{\bm{e}_p^i, i\in \mathcal{M}\}\in\mathbb{R}^{|\mathcal{M}|\times D}$ is a learnable embedding for the masked patches. The output of the encoder is $\{\bm{h}_i\}_{i=1}^P$ ($\bm{h}_i\in\mathbb{R}^D$) represents the encoded representations of the $i$-th patch.

 \item For MAE, the ViT encoder takes only the linear projection of the visible patches $\bm{x}_p^\mathcal{V}\bm{E}$ as the input with added position embeddings. The output of the encoder is $\{\bm{h}_i, i\in\mathcal{V}\}$ where $\bm{h}_i\in\mathbb{R}^D$ represents the encoded visible patch $i\in\mathcal{V}$.
\end{itemize}

\subsubsection{Decoder} 
Our decoder performs the signal reconstruction task given the encoded representations of the input patches.
 
\begin{itemize}
    \item For BEiT, the inputs to the decoder are the encoded representations for all the patches $\{\bm{h}_i\}_{i=1}^P$ obtained from the last layer of the encoder. The decoder is a single linear layer to predict the visual tokens at the masked positions $\{\bm{z_i}, i\in\mathcal{M}\}$ which was generated by the DALLE pre-trained dVAE\cite{ramesh2021zero} tokenizer. 
    \item For MAE, the inputs to the decoder are the encoded visible patches $\{\bm{h}_i, i\in\mathcal{V}\}$ along with a learnable vector for the masked patches $\bm{e}_p^\mathcal{M}=\{\bm{e}_p^i, i\in \mathcal{M}\}$ and position embeddings. The decoder is a lightweight ViT that regresses the pixel values for the masked patches.
\end{itemize}

%==================Added algorithm===========
\setlength\textfloatsep{0.4\baselineskip}
\SetKwBlock{ClientExecute}{ClientUpdate($k$, $w_t$)}{}
\SetKwBlock{ServerExecute}{Server Execution:}{}
\SetKwComment{Comment}{$\triangleright$}{}
% \SetAlgoSkip{}

\LinesNumberedHidden{
\begin{algorithm}[t]
	\caption{Our generalized self-supervised FL framework. $T$ is the maximum number of communication rounds, $E$ is the number of local epochs.}
	\label{algo:FL_pretraining}
\KwIn{local client $k$, local data $\mathcal{D}^k=\mathcal{D}^k_l\bigcup\mathcal{D}^k_u$}

\ServerExecute{
initialize $w_0$~\\
\For{each round $t=1,...,T$}
{
$S_t$ $\leftarrow$ (\text{Selection of $K$ clients})~\\
\For{each client $k\in S_t$ in parallel}
{
$w^k_{t+1}$ $\leftarrow$ \text{ClientUpdate}($k$, $w_t$)~\\
}
\Comment{Pre-training stage}
$w_{t+1} \leftarrow \sum_{k=1}^K\frac{|\mathcal{D}^k|}{|\bigcup\mathcal{D}^k|}w_{t+1}^k$~\\
\medskip

\Comment{Fine-tuning stage}
$w_{t+1} \leftarrow \sum_{k=1}^K\frac{|\mathcal{D}_l^k|}{|\bigcup\mathcal{D}_l^k|}w_{t+1}^k$~\\
}
}
\medskip

\ClientExecute{
\Comment{Pre-training stage}
Sample batches $\mathcal{B}$ from local data $\mathcal{D}^k$ \\
\For{each local epoch $i=1,...,E$}
{
\For{batch $b \in \mathcal{B}$}
{
$b^{\mathcal{M}} = \bigcup_{\bm{x}\in b}\text{Masking}(\bm{x})=\bigcup_{\bm{x}\in b}\bm{x}^\mathcal{M}$ 

$w_{t+1}^k \leftarrow w_{t,k} - \eta\nabla\ell_k(w_{t,k}, b^{\mathcal{M}})$
}
}
\Comment{Fine-tuning stage}
Sample batches $\mathcal{B}$ from local labeled data $\mathcal{D}^k_l$ \\
\For{each local epoch $i=1,...,E$}
{
\For{batch $b \in \mathcal{B}$}
{
$w_{t+1}^k \leftarrow w_{t,k} - \eta\nabla\ell_k(w_{t,k}, b)$ \\
}
}
}
\end{algorithm}
}
%==========================end of algorithm

\subsubsection{Loss function}
The $k$-th local encoder $E_k$ and decoder $D_k$ are trained with their local data $\mathcal{D}^k$ to minimize the local objective function $\mathcal{L}_k(w)=\mathbb{E}_{x\sim\mathcal{D}^k}[\ell_k(w;\bm{x})]$.
\begin{itemize}
    \item In BEiT, $\ell_k$ is the cross-entropy loss of the predicted visual tokens of the masked patches $\{\bm{z}_i, i\in \mathcal{M}\}\in\mathbb{R}$: 
    \begin{equation}
\ell_k =-\sum_{i\in\mathcal{M}}\frac{1}{|\mathcal{M}|}\log p(\bm{z}_i;w|\bm{x}^\mathcal{M}),
\end{equation}
    
    \item In MAE, $\ell_k$ is the mean squared error of the predicted pixel values of the masked patches $\{\bm{x}_p^i, i\in\mathcal{M}\}\in\mathbb{R}^{S\times S\times C}$:
\begin{equation}
\ell_k=\sum_{i\in\mathcal{M}}\frac{1}{|\mathcal{M}|}((\bm{x}_p^i - \hat{\bm{x}}_p^i)^2; w),
\end{equation}
\end{itemize}

In the federated pre-training stage, each local client takes $E$ steps of gradient descent to update the local model $E_k$ and $D_k$ by minimizing its local loss $\mathcal{L}_k$ on data $\mathcal{D}^k$. 
Then, the server takes a weighted average of all the resulting local models to update the global model $E_G$ and $D_G$, which is further sent back to the local clients for the next training iteration. The whole pre-training process terminates when we reach the maximum number of communication rounds $T$. Once pre-training is complete, the final pre-trained global encoder ($E_G^*$) is saved while the final global decoder ($D_G^*$) is discarded.

%+++++++++++++++++++++++ 3-4: Supervised Federated Fine-tuning
\subsection{Supervised Federated Fine-tuning}
\label{sec:supervised finetuning}
In the federated fine-tuning stage, as shown in Fig.~\ref{Fig:framework}, 
we initialize the local encoder $E_k$ of the $k$-th client with the pre-trained global encoder $E_G^*$ obtained from the first stage, and append a linear classifier $L_k$ upon the encoder.
The entire model is then fine-tuned on the local labeled data. Specifically, we use average pooling to extract the learned representations from the local encoder, which are then fed to a linear classifier $L_k$ (\emph{i.e.}, a softmax layer) to minimize the cross-entropy loss for image classification tasks.

%============================= SECTION 4: EXPERIMENTS =============================%

\section{Experiments}
In this section, we present experiments on a variety of simulated and real-world federated medical datasets to evaluate the effectiveness of our methods.

We first provide details on the datasets and experimental setup, then compare the robustness of our methods to data heterogeneity with baselines that are pre-trained on ImageNet.
Additionally, we analyze the generalizability of the proposed method to out-of-distribution data and investigate its label efficiency through fine-tuning with different fractions of labels.
Furthermore, we compare the performance of our methods to previous FL methods, including (1) federated self-supervised pre-training baselines and (2) optimization-based FL methods and semi-supervised FL methods, in terms of robustness to non-IID data and label efficiency.

%+++++++++++++++++++++++ 4-1: DATASET
\vspace{-.5em}
\subsection{Dataset}
\begin{figure}[t]
% \vspace{-0.5em}
\begin{center}
    \includegraphics[width=\linewidth]{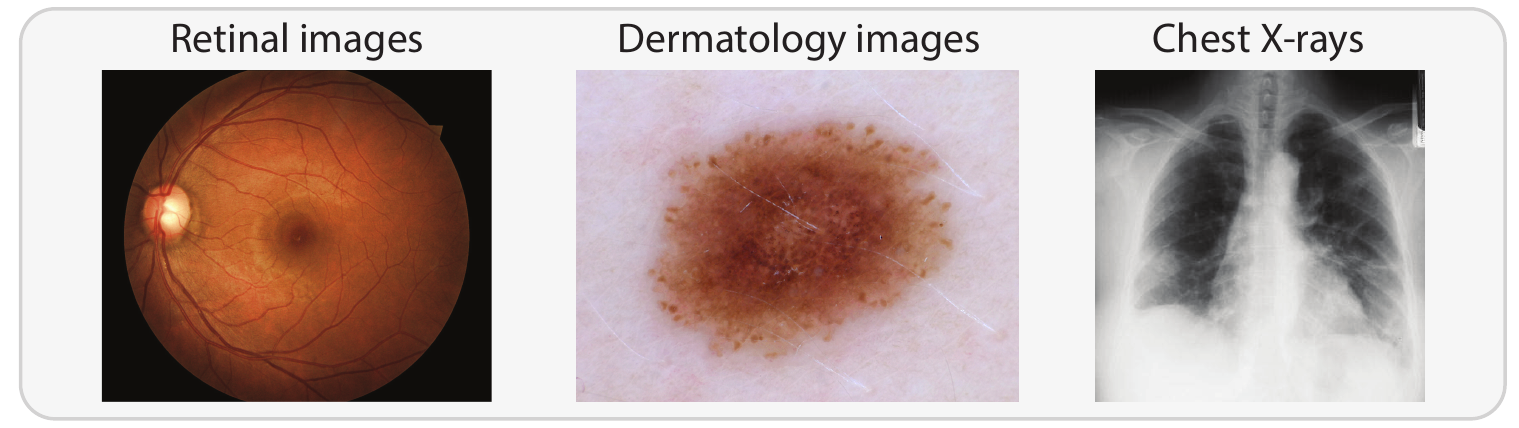}
\end{center}
\vspace{-1.2em}
\caption{Preview of retinal images, skin images and chest X-rays} 
\label{Fig:samples}
\vspace{-.2em}
\end{figure}

We evaluate the performance of our methods on three popular tasks in the medical imaging domain: (1) detecting diabetic retinopathy from retinal fundus images, (2) diagnosing skin lesions from dermatology images, and (3) identifying pneumonia and COVID-19 from chest X-rays. These tasks vary in terms of image modality, image acquisition, label distribution, and other factors. 
For example, retinal images are obtained using fundus cameras, dermatology images are captured with digital cameras, and chest X-rays are acquired using X-ray scanners. Fig.~\ref{Fig:samples} illustrates the visual differences among these three medical datasets.

%----------fig3 start (Samples for preview)--------
%----------fig3 end --------

% RETINA DATASET
\vspace{0.5ex}\noindent\textbf{Retina Dataset.} 
We evaluate FL methods on the Kaggle Diabetic Retinopathy competition dataset\footnote{https://www.kaggle.com/c/diabetic-retinopathy-detection}, which contains 35,126 retinal fundus images acquired from various cameras. 
The original images are divided into five categories (normal, mild, moderate, severe, and proliferating). 
We preprocessed the dataset by binarizing the labels into Normal and Diseased, and randomly selecting 9,000 balanced images as the training set and 3,000 images as the test set. 

% DERM DATASET
\vspace{0.5ex}\noindent\textbf{Dermatology Dataset.} This dermatology dataset is referred to as \emph{Derm} in this paper. 
It includes images from ISIC17, 19, 20\footnote{https://challenge.isic-archive.com/data/}, with approximately 5,000 images in the Melanoma (malignant) class from these three datasets and 5,000 randomly selected images in the benign classes from ISIC19. 
The Derm dataset is then randomly divided into a training set of approximately 7,500 images and a test set of approximately 2,500 images.

% COVID-FL DATASET
\vspace{0.5ex}\noindent\textbf{COVID-FL dataset.}
To evaluate the performance of our methods on a real-world federated data partitions, we create \emph{COVID-FL}, a dataset in which each client only contains data from a single real-world site without any overlap.
Our COVID-FL dataset includes 20,018 chest X-ray scans from eight different publicly available data repositories: 
(1) BIMCV-COVID19~\cite{vaya2020bimcv} (the Valencia Region Image Bank, Spain),
(2) ml-workgroup\footnote{https://github.com/ml-workgroup/covid-19-image-repository} (the Institute for Diagnostic and Interventional Radiology, Hannover, Germany),
(3) SIRM\footnote{https://www.sirm.org/category/senza-categoria/COVID-19/} (the Italian Society of Medical and interventional Radiology COVID-19 Database, Italy), 
(4) Eurorad\footnote{https://eurorad.org/} (the European Society of Radiology),
(5) MIDRC-RICORD-1c\footnote{https://doi.org/10.7937/91ah-v663} (the RSNA International COVID-19 Open Radiology Database),
(6) the RSNA Pneumonia Detection Challenge dataset\footnote{https://www.kaggle.com/c/rsna-pneumonia-detection-challenge/},
(7) the Guangzhou pediatric dataset~\cite{kermany2018identifying} (from the Guangzhou Women and Children’s Medical Center, China), and
(8) the Cohen dataset\footnote{https://github.com/ieee8023/covid-chestxray-dataset} with duplicated images removed. 

In our COVID-FL dataset, each data site represents a single medical institution in order to mimic the real-world federated scenarios.
Each site may be missing one or more classes. 
For example, BIMCV only contains images of COVID-19 infections, while Guangzhou pediatric only includes images of normal and non-COVID-19 pneumonia patients (Fig.~\ref{6a}). 
Data from different sites were acquired using different machines and on different patient populations, resulting in heterogeneity in intensity distribution (Fig.~\ref{6b}). 
This simulates the real-world scenario where data is collected from various institutions with different equipment and patient populations.

The COVID-FL dataset is further divided into an 80\%-20\% train-test split, yielding 16,044 training images and 3,974 test images. 
Each data site has the same proportion of train and test sets. 
The test set can be considered as a combination of the held-out data in each client (\emph{i.e.}, hospital).

% SKIN-FL DATASET
\vspace{0.5ex}\noindent\textbf{Skin-FL dataset.} 
The Skin-FL dataset contains skin lesion images and was created to evaluate the generalization of the model to out-of-distribution data. 
Following \cite{bdair2021fedperl}, after removing duplicates, the training set of Skin-FL consists of 22,888 images from four datasets as shown in Fig.~\ref{Fig:skin_dist}a: 784 images from Derm7pt~\cite{kawahara2018seven}, 8,012 images from HAM10000~\cite{tschandl2018ham10000}, 1,839 images from PAD-UFES~\cite{pacheco2020impact}, and 12,253 images from ISIC19. 
There are eight classes in total in the training set: Actinic keratosis (AK), Benign keratosis (BKL), Melanoma (MEL), Melanocytic nevus (NV), Vascular lesion (VASC), Squamous cell carcinoma (SCC), Basal cell carcinoma (BCC), and Dermatofibroma (DF). 
Specifically, Derm7pt has six classes except for AK and SCC; HAM10000 includes seven classes except for SCC; PAD-UFES contains six classes except for DF and VASC; ISIC19 contains all eight classes. 

Moreover, we use 33,126 images from ISIC20~\cite{rotemberg2021patient} as our out-of-distribution test set to investigate how our proposed method generalizes to unseen clients. 
This test set contains several classes that are not included in the training set, so we binarize the predictions during fine-tuning and the labels of our test set into Benign and Malignant following~\cite{bdair2021fedperl}.
This task is very challenging due to the severe class imbalance (Fig.~\ref{7b}).

%+++++++++++++++++++++++ 4-2: EXPERIMENT SETUP
\subsection{Experiment Setup}
% Construction of non-IID dataset
\subsubsection{Construction of non-IID dataset}
We model IID and non-IID data distributions using a Dirichlet distribution following \cite{yurochkin2019bayesian, hsu2019measuring, zhu2021data} for Retina and Derm
dataset. 
Compared to real federated data partitions, simulated data partitions allow for a more flexible and thorough investigation of the model behavior, as they can be easily manipulated to test different degrees of data heterogeneity. Suppose a dataset has $J$ classes, we randomly partition the data into $N$ local clients by simulating $$\bm{p}_j=\{p_{j,1}, ..., p_{j,N}\}\sim \text{Dir}_N(\alpha)$$
where $p_{j,k}\in(0,1)$ and $||\bm{p}_j||_1=1$ ($j\in[1,J]$, $k\in[1,N]$). 

\begin{figure}[t]
\begin{center}
    \includegraphics[width=\linewidth]{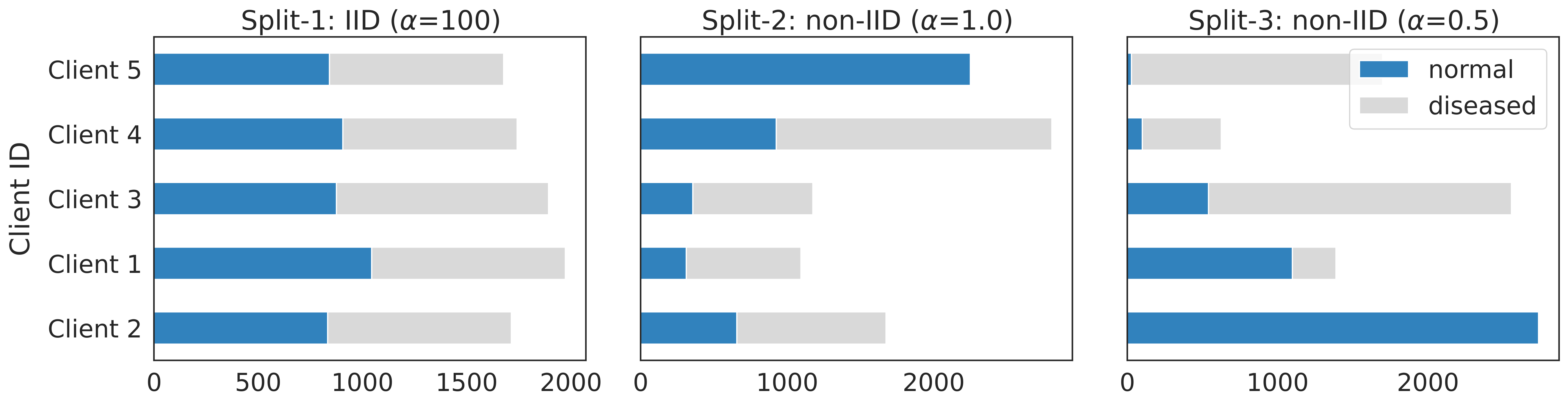}
\end{center}
\vspace{-.8em}
\caption{Retina: visualization of statistical heterogeneity among clients, where the $x$-axis is the number of training samples, and the $y$-axis is client IDs.} 
\label{Fig:retina_dist}
\vspace{-0.5em}
\end{figure}

\begin{figure}[t]
\begin{center}
    \includegraphics[width=\linewidth]{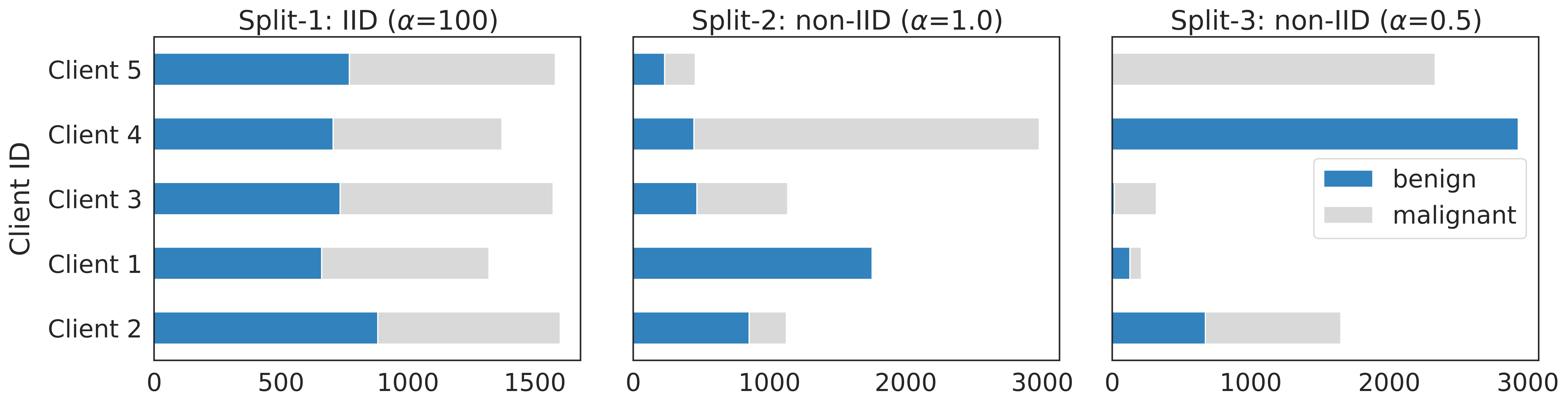}
\end{center}
\vspace{-.8em}
\caption{Derm: visualization of statistical heterogeneity.} 
\label{Fig:derm_dist}
\end{figure}

We assign a proportion $p_{j,k}$ of the instances of class $j$ to client $k$. 
The concentration parameter $\alpha$ in the Dirichlet distribution $\text{Dir}(\alpha)$ controls the degree of heterogeneity, with smaller values of $\alpha$ leading to higher data heterogeneity. 
We simulate three sets of data partitions with $\alpha$ values of ($\alpha=\{100,1.0,0.5\}$) for both the Retina and Derm datasets, each of which consists of $N=5$ simulated clients (see Fig.~\ref{Fig:retina_dist} and \ref{Fig:derm_dist}).
Based on the level of data heterogeneity, these three partitions are referred to as Split-1 (IID), Split-2 (moderate non-IID) and Split-3 (severe non-IID).

\begin{figure}[t]
\begin{center}
  \subfloat[Label distribution on COVID-FL\label{6a}]{%
      \includegraphics[width=0.53\linewidth]{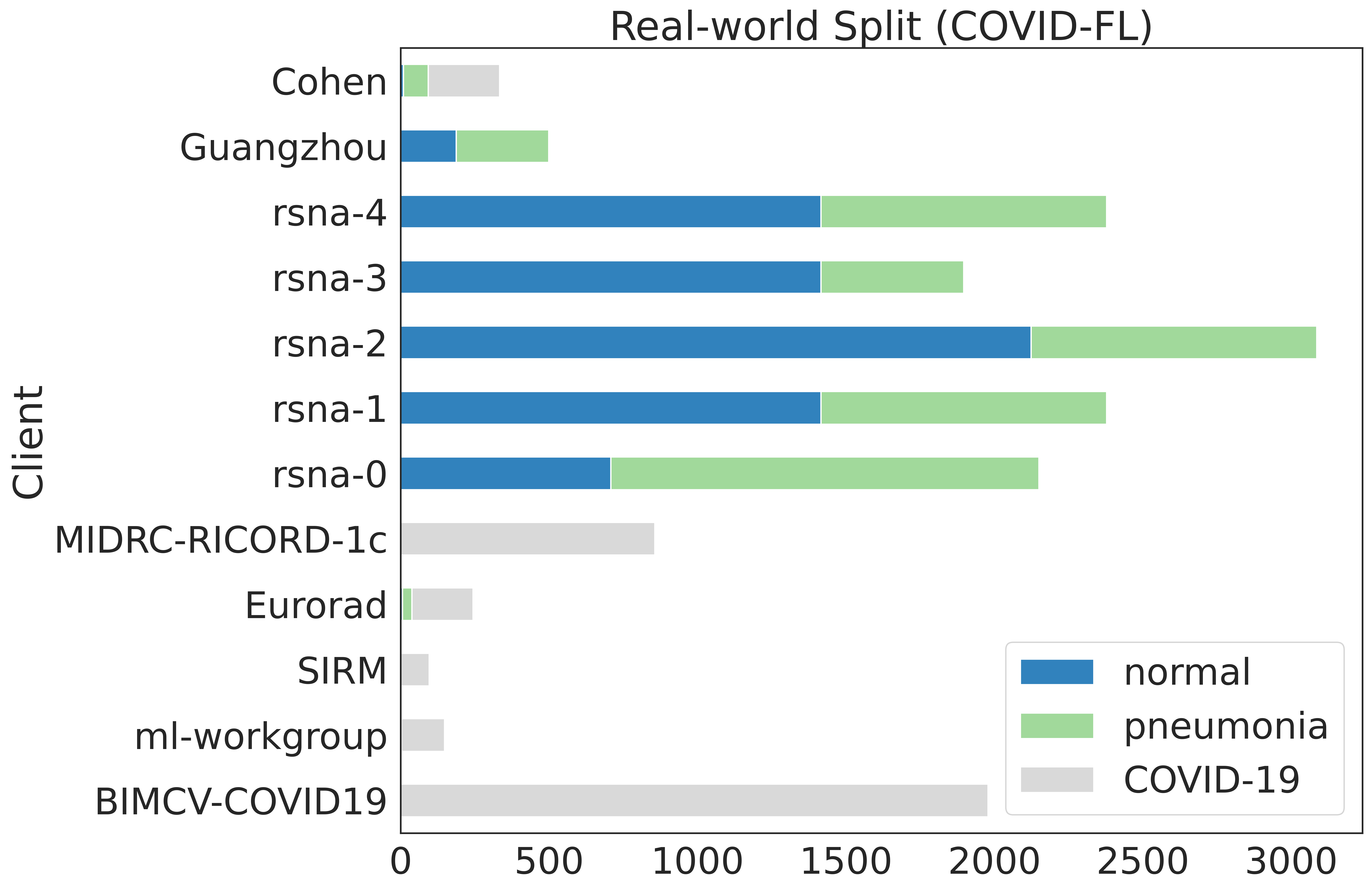}}
  \hfill
  \subfloat[Chest X-ray intensity distribution at each client\label{6b}]{%
      \includegraphics[width=0.44\linewidth]{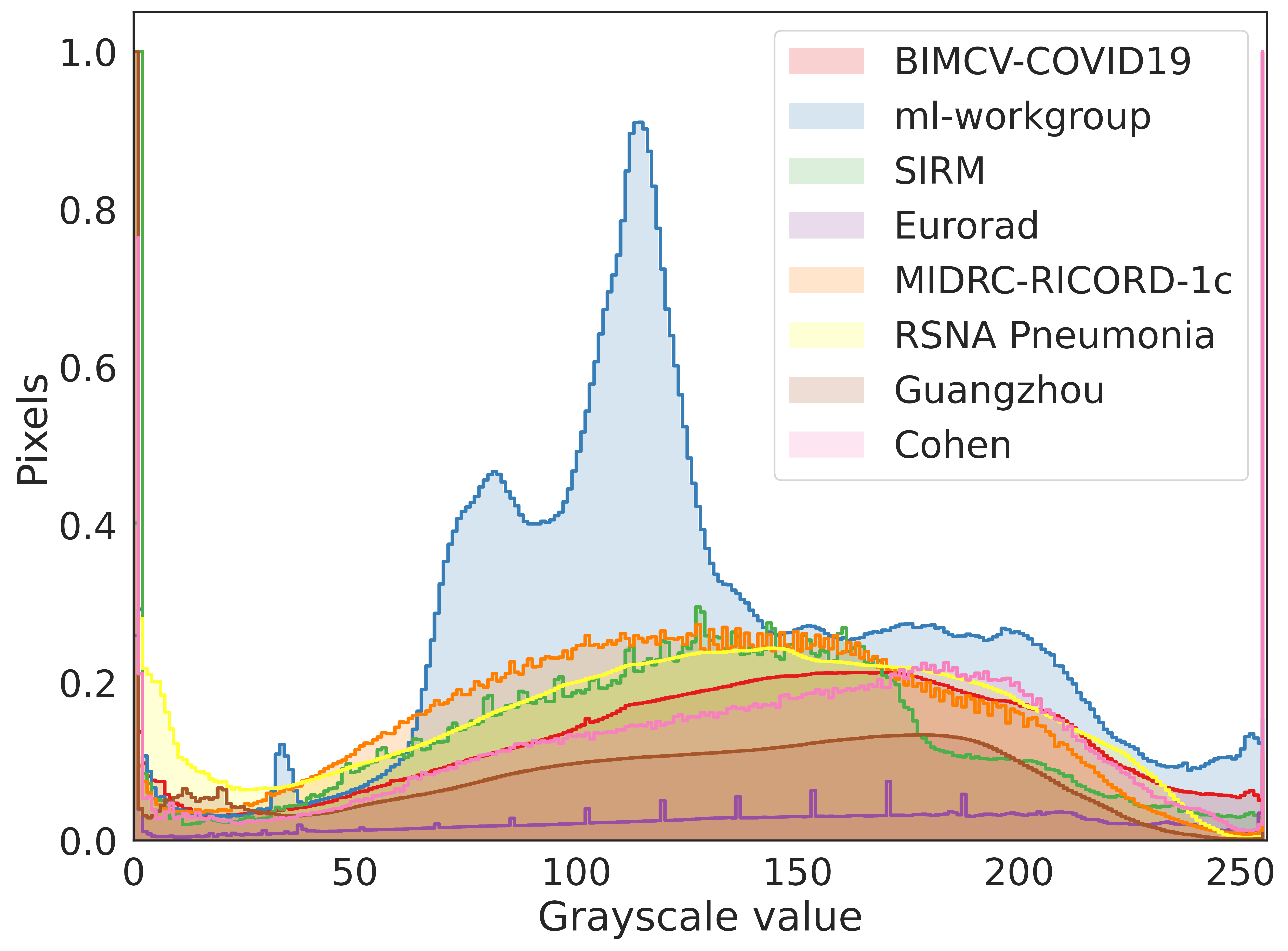}}
\end{center}
\vspace{-1em}
\caption{COVID-FL: data heterogeneity among clients.} 
\label{Fig:covid_dist}
\vspace{-1.5em}
\end{figure}

\begin{figure}[t]
\begin{center}
  \subfloat[Label distribution on Skin-FL train set\label{7a}]{%
      \includegraphics[width=0.55\linewidth]{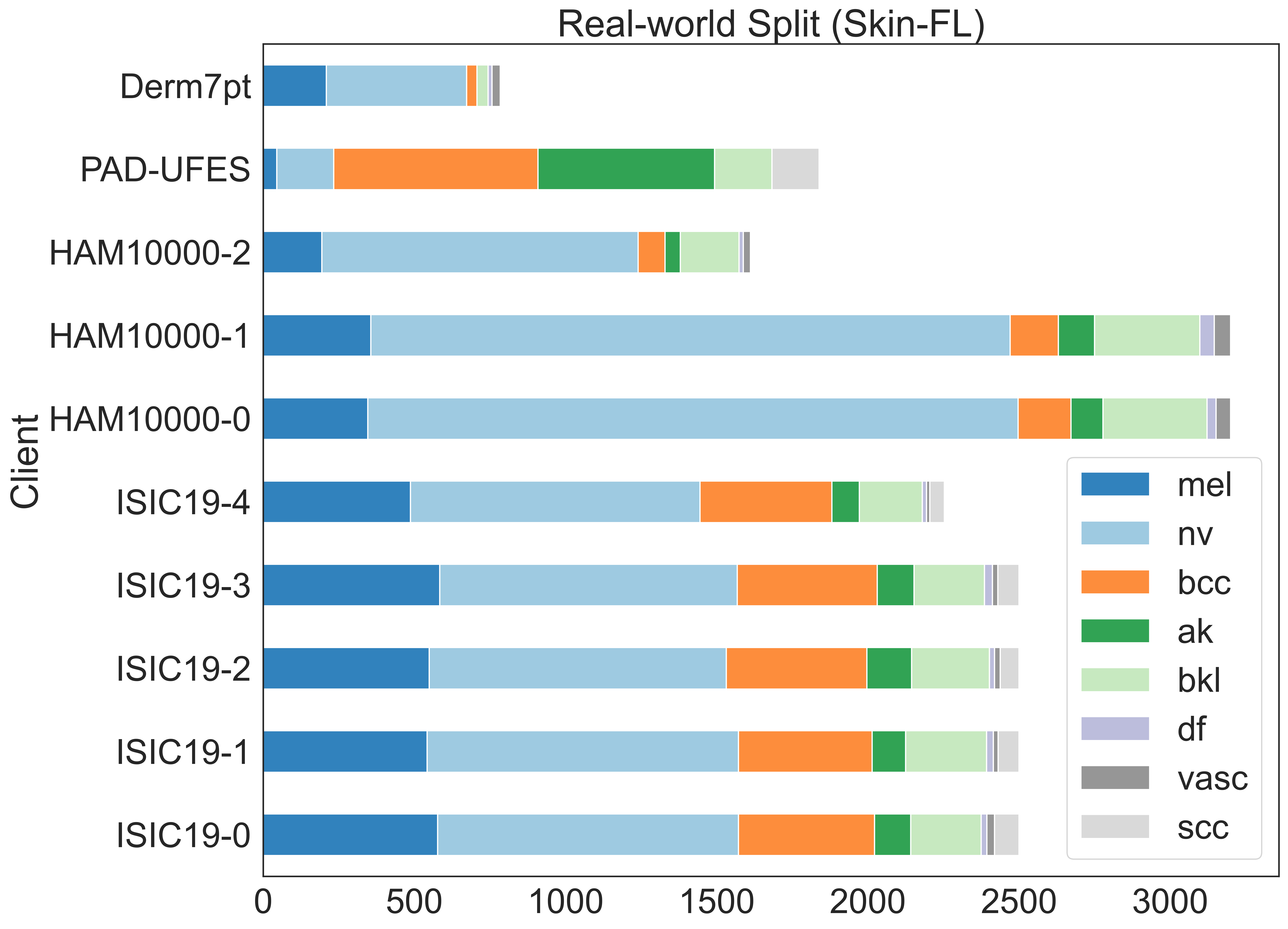}}
  \hfill
  \subfloat[Out-of-distribution test set\label{7b}]{%
     \includegraphics[width=0.42\linewidth]{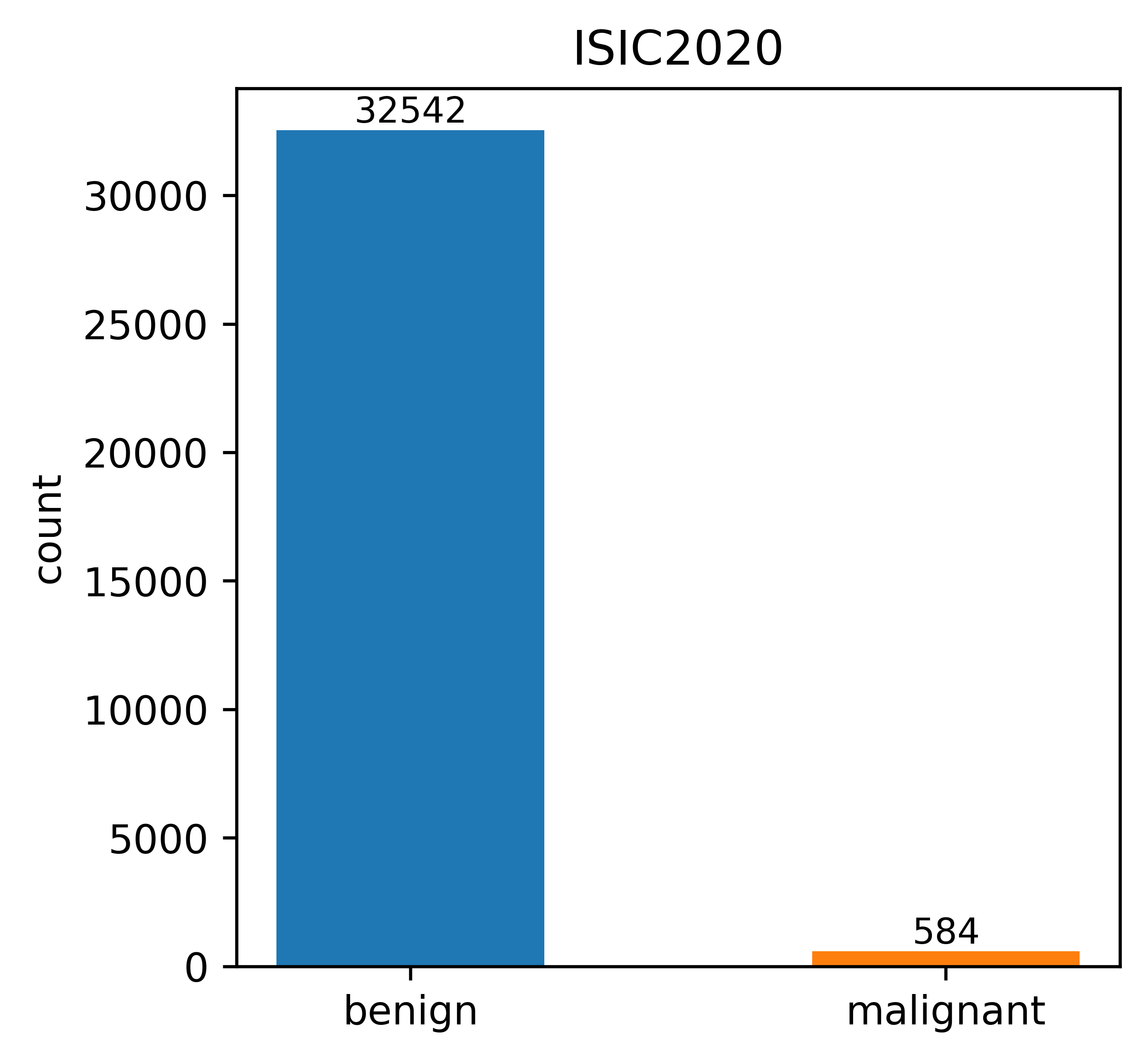}}
\end{center}
\vspace{-1em}
\caption{Skin-FL: data heterogeneity among clients.} 
\label{Fig:skin_dist}
\end{figure}

COVID-FL and Skin-FL are two real-world federated data sets that exhibit both label distribution skewness and feature distribution skewness.
Given that some clients contain significantly more data than others, 
we partition those clients into sub-clients without any overlap, resulting in a total of 12 clients in COVID-FL and 10 clients in Skin-FL. 
The distributions of these two datasets are shown in Fig.~\ref{6a} and~\ref{7a}.

%----------fig4 start (RETINA DISTRIBUTION)--------
%----------fig4 end --------

%----------fig5 start (DERM DISTRIBUTION)--------
%----------fig5 end --------

%----------fig6 start (COVID-FL DISTRIBUTION)--------
%----------fig6 end --------

%----------fig7 start (SKIN-FL DISTRIBUTION)--------
%----------fig7 end --------

% DATA AUGMENTATION
\subsubsection{Data Augmentation}
During pre-training, we apply random scaling and crop patches of size $224\times224$ from the original images for all datasets, followed by random color jittering and random horizontal flipping. 
The random scaling factor is chosen from a range of [0.4, 1.0] for COVID-FL and [0.2, 1.0] for the other datasets. 
During fine-tuning, we apply random scaling and cropping the images patches of size $224\times224$ and perform random rotation with a degree of 10 and random horizontal flipping. 
The random scaling factor is chosen from a range of [0.8, 1.2] for COVID-FL and [0.6, 1.0] for the other datasets.

% SELF-SUPERVISION FL PRE-TRAINING SETUP
\subsubsection{Self-supervised FL pre-training setup}
All methods are implemented using Pytorch and deployed in a distributed training system using DistributedDataParallel (DDP). 
ViT-B \cite{dosovitskiy2021an} is chosen as the backbone for the proposed models. 
Following the setup in BEiT\cite{bao2021beit} and MAE\cite{he2021masked}, the input is split into $14\times14$ image patches and the same number of visual tokens for BEiT and $16\times16$ patches for MAE.
In our main experiment, we randomly mask at most $40\%$ of total image patches for BEiT and $60\%$ for MAE.
AdamW with $\beta1 = 0.9$, $\beta2 = 0.999$ is used for optimization.

We use the same set of hyperparameters for both centralized and federated learning in each task. 
The base learning rate ($\eta$) and batch size ($B$) vary among tasks based on hyperparameter tuning. 
More details can be found in Table~\ref{tab:hyperparam}. 
Fed-BEiT pre-training runs for 1000 communication rounds with a warmup period of 10 epochs; Fed-MAE pre-training runs for 1600 communication rounds with a warmup period of 5 epochs. Both methods employ a cosine learning rate decay of 0.05.

In terms of the pre-training schedule, we observe that a larger number of communication rounds generally leads to more improvement, but the improvement becomes much less prominent after a certain number of rounds.
Here, we use 1000 and 1600 communication rounds as the default values for Fed-BEiT and Fed-MAE pre-training, respectively. 
Note that we 
 also conduct ablation studies on the number of communication rounds and investigate its impact on model performance in Sec.~\ref{sec:ablation}.

%----------table1 start (hyper-parameters)--------
\begin{table}[t]
\footnotesize
\centering
\caption{Table of hyper-parameters (base learning rate $\eta$ and batch size $B$) in experiments on the Retina, Derm, COVID-FL and Skin-FL datasets during federated self-supervised pre-training.}
\label{tab:hyperparam}
\vspace{-0.5em}
\begin{tabular}{cccccc}
\toprule[0.15em]
\multicolumn{1}{c}{\multirow{2}{*}{Dataset}}  
&\multicolumn{2}{c}{Fed-BEiT}  
& \multicolumn{2}{c}{Fed-MAE} \\
\cline{2-3} \cline{4-5}
\multicolumn{1}{c}{} \T & $\eta$ & $B$ &$\eta$ & $B$ \B\\
\midrule
Retina & 1.5e-3 & 256 & 1e-3 & 128 \\
Derm & 3e-3 & 128 & 7.5e-4 & 128  \\
COVID-FL &1.5e-3 & 64 & 3.75e-4 & 64 \\
Skin-FL &1.5e-3 & 128 & 7.5e-4 & 128 \\
\bottomrule[0.15em]
\end{tabular}
\vspace{-1em}
\end{table} 
%----------table1 end --------

%SUPERVISED FL FINE-TUNING FOR DOWNSTREAM TASKS
\subsubsection{Supervised FL fine-tuning for downstream tasks}
During federated fine-tuning, the model is fine-tuned for 100 communication rounds with a base learning rate starting at 3e-3 and a batch size of 256 for all tasks, except for COVID-FL, whose batch size is set to 64.

%EVALUATION METRICS
\subsubsection{Evaluation Metrics}
We use accuracy as the evaluation metric for classification on the Retina, Derm and COVID-FL datasets. 
For the Skin-FL dataset, due to its severe class imbalance issues, we use F1-score as the evaluation metric.

%+++++++++++++++++++++++ 4-3: MAIN RESULTS: MODEL PERFORMANCE 
\subsection{Results}
% %----------fig8 start (RESULT COMPARISON IN PLOT)--------
\begin{figure}[t]
\begin{center}
    \includegraphics[width=.96\linewidth]{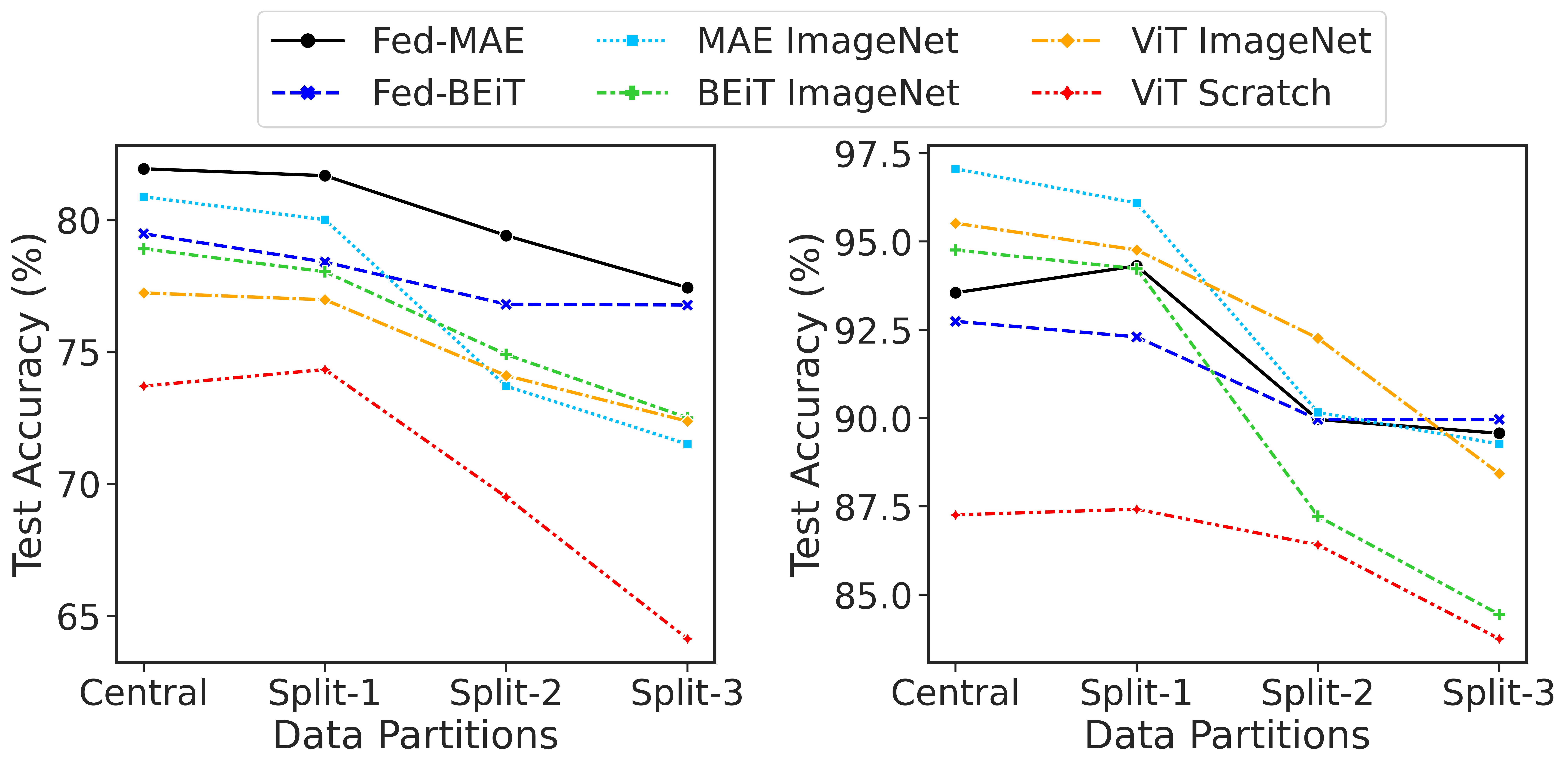} 
    \vskip -1.7em
    \subfloat[\label{8a} Results on Retina]{\hspace{.48\linewidth}}
    \subfloat[\label{8b} Results on Derm]{\hspace{.48\linewidth}}
    \vspace{.3em}
    \includegraphics[width=.96\linewidth]{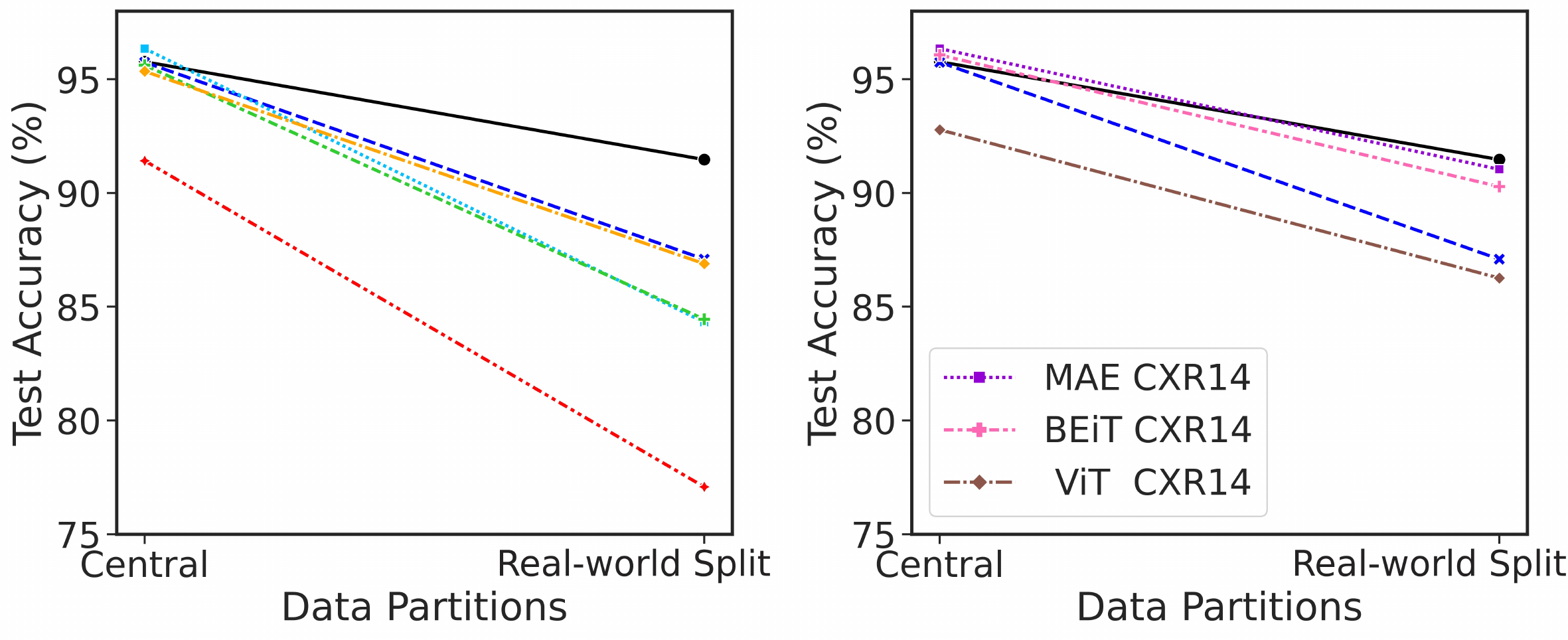}
    \vskip -1.7em
    \subfloat[\label{8c} Results on COVID-FL (left: comparison with training from scratch and with ImageNet pre-training; right: comparison with CXR14 pre-training)]{\hspace{\linewidth}}
 \end{center}
 \vspace{-.8em}
 \caption{Comparison of model performance in terms of test accuracy w.r.t. data heterogeneity for the Retina, Derm and COVID-FL datasets. 
 }
\label{Fig:main_results} 
\end{figure}
% %----------fig8 end (RESULT COMPARISON IN PLOT)--------

To evaluate the proposed federated self-supervised pre-training methods (Fed-BEiT and Fed-MAE), we compare them with four baseline approaches, including: (1) no pre-training (ViT scratch), (2) ImageNet supervised pre-training (ViT ImageNet)~\cite{qu2021rethinking}, (3) ImageNet pre-training using BEiT~\cite{bao2021beit} (BEiT ImageNet), and (4) ImageNet pre-training using MAE~\cite{he2021masked} (MAE ImageNet).

We pre-train methods directly on decentralized target task data in a distributed setting, while the four baseline approaches (except for ViT Scratch which does not require pre-training) are pre-trained on a centralized large-scale dataset ImageNet-22K~\cite{deng2009imagenet} under centralized settings.
Furthermore, on COVID-FL, we conduct experiments using pre-trained models trained on the large dataset ChestX-ray14 (CXR14)~\cite{wang2017chestx} as additional baselines (ViT CXR14, BEiT CXR14 and MAE CXR14), similar to the three ImageNet pre-training baselines.
CXR14 consists of 112,120 chest X-ray images, which is seven times larger in size than the COVID-FL training set.

%----------table2 start --------
\begin{table}[t]
\footnotesize
\centering
\caption{Test accuracy for federated fine-tuning on Retina Central (centralized), Split-1 (IID), Split-2 (moderate non-IID) and Split-3 (severe non-IID). The best result is bolded and the second-best result is marked with a line underneath.}
\vspace{-0.5em}
\label{tab:retina}
\begin{tabularx}{0.48\textwidth}{@{\extracolsep{4pt}}s e d f f f f}
\toprule[0.15em]
\multicolumn{3}{c}{Pre-training}
&\multicolumn{4}{c}{Test Accuracy (\%)}
\B
\\
\cline{1-3} \cline{4-7}
\T Method
&Setup
&Dataset
&Central
&Split1
&Split2
&Split3
\\
\midrule
\emph{None}
&\emph{None}
&\emph{None}
&73.70 &74.33 &69.50 &64.13
\\
Supervised
&Centralized
&ImageNet
&77.23 &76.97 &74.10 &72.37 
\\
BEiT~\cite{bao2021beit}
&Centralized
&ImageNet
&78.90 &78.03 &74.90 &72.50 
\\
MAE~\cite{he2021masked}
&Centralized
&ImageNet
&\underline{80.87} &\underline{80.00} &73.70 &71.50
\\
\midrule
Fed-BEiT
&Distributed
&Retina
&79.47 &78.40 &\underline{76.80} &\underline{76.77}
\\
Fed-MAE
&Distributed
&Retina
&\textbf{81.93} &\textbf{81.67} &\textbf{79.40} &\textbf{77.43}\\
\bottomrule[0.15em]
\end{tabularx}
\vspace{-1em}
\end{table} 
%----------table2 end --------

%----------table3 start --------
\begin{table}[t]
\footnotesize
\centering
\caption{Test accuracy for federated fine-tuning on the Derm dataset with various degrees of data heterogeneity.}
\vspace{-0.5em}
\label{tab:derm}
\begin{tabularx}{0.48\textwidth}{@{\extracolsep{4pt}}s e d f f f f}
\toprule[0.15em]
\multicolumn{3}{c}{Pre-training}
&\multicolumn{4}{c}{Test Accuracy (\%)}
\B
\\
\cline{1-3} \cline{4-7}
\T Method
&Setup
&Dataset
&Central
&Split1
&Split2
&Split3
\\
\midrule
\emph{None}
&\emph{None}
&\emph{None}
&87.26 &87.42 &86.41 &83.75
\\
Supervised
&Centralized
&ImageNet
&\underline{95.52} &\underline{94.76} &\textbf{92.26} &88.43
\\
BEiT~\cite{bao2021beit}
&Centralized
&ImageNet
&94.76 &94.23 &87.22 &84.44 
\\
MAE~\cite{he2021masked}
&Centralized
&ImageNet
&\textbf{97.06} &\textbf{96.09} &\underline{90.16} &89.27
\\\midrule
Fed-BEiT
&Distributed
&Derm
&92.74 &92.30 &89.96 &\textbf{89.96}
\\
Fed-MAE
&Distributed
&Derm
&93.55 &94.31 &89.96 &\underline{89.57}\\
\bottomrule[0.15em]
\end{tabularx}
\end{table} 
\vspace{-.5em}
%----------table3 end --------

All of the methods in this comparison utilize ViT-B~\cite{dosovitskiy2021an} as their backbone for fairness.
The federated fine-tuning process is run for 1000 communication rounds for models trained from random initialization, and 100 rounds for the others.

\subsubsection{More robust to data heterogeneity} 
Data heterogeneity is a key FL challenge that our work aims to address. 
Fig.~\ref{Fig:main_results} and Table~\ref{tab:retina}-\ref{tab:COVID-FL} compare the results of our proposed method and the baselines under different degrees of data heterogeneity for multiple medical image classification tasks.

First, we observe that our proposed method is the only method that is consistently robust across all medical tasks under different levels of data heterogeneity. 
Specifically, the discrepancy in test accuracy across different data partitions is smallest using our methods, Fed-BEiT and Fed-MAE. The advantage of our method is particularly pronounced when data heterogeneity is severe. 
In particular, our methods outperform the four baselines when the data distribution is strongly skewed (\emph{i.e.}, Retina and Derm Split-3 and COVID-FL split), with an improvement of 5.06\%, 1.53\%, and 4.58\% in test accuracy on Retina, Derm and COVID-FL, respectively, compared to the supervised baseline with ImageNet pre-training. 
While our methods outperform all the baselines in severe non-IID scenarios, it is worth noting that they may not always outperform under centralized and IID settings for certain tasks, such as Derm. 
This is reasonable, as the domain shift between the pre-training dataset ImageNet and the fine-tuning skin images is relatively small in this case. 
Nonetheless, our method consistently improves model performance under severe data heterogeneity across these three medical tasks.

%----------table4 start --------
\begin{table}[t]
\footnotesize
\centering
\caption{Test accuracy for federated fine-tuning on COVID-FL Central (centralized) and Split (non-IID). $T$ represents the communication rounds of federated fine-tuning.}
\vspace{-0.5em}
\label{tab:COVID-FL}
\begin{tabularx}{0.48\textwidth}{@{\extracolsep{2pt}}e d g h h f}
\toprule[0.15em]
\multicolumn{3}{c}{\centering Pre-training}
&\multicolumn{2}{c}{\centering Test Accuracy (\%)}
&\multicolumn{1}{f}{\multirow{2}{*}{$T$}}
\B
\\
\cline{1-3} \cline{4-5}
\T Method
&Setup
&Dataset
&Central
&Split
\\
\midrule
\emph{None}
&\emph{None}
&\emph{None}
&91.42
&77.08
&1000
\\
Supervised
&Centralized
&ImageNet
&95.35 &86.89 &100
\\
BEiT~\cite{bao2021beit}
&Centralized
&ImageNet
&95.62 &84.45 &100
\\
MAE~\cite{he2021masked}
&Centralized
&ImageNet
&\textbf{96.35} &84.32 &100
\\\midrule
Supervised
&Centralized
&ChestX-ray14
&92.78 &82.26 &100
\\
BEiT~\cite{bao2021beit}
&Centralized
&ChestX-ray14
&\underline{96.07} &90.28 &100
\\
MAE~\cite{he2021masked}
&Centralized
&ChestX-ray14
&\textbf{96.35}
&\underline{91.04}
&100
\\
\midrule
Fed-BEiT
&Distributed
&COVID-FL
&95.75
&87.09
&100
\\
Fed-MAE
&Distributed
&COVID-FL
&95.77
&\textbf{91.47}
&100
\\
\bottomrule[0.15em]
\end{tabularx}
\end{table} 
%----------table4 end --------

We further investigate the performance of ImageNet self-supervised pre-training methods, including BEiT ImageNet and MAE ImageNet, which have been shown to outperform their supervised counterparts when fine-tuning with ImageNet in centralized learning~\cite{bao2021beit}\cite{he2021masked}. 
However, we find that these methods are more prone to non-IID data compared to their supervised counterparts and our federated self-supervised learning methods, resulting in a nontrivial decline in performance when the label distribution skewness among clients increases.

We have demonstrated that our proposed method, which conducts federated self-supervised pre-training using only the decentralized target task medical images (with much smaller size compared to other pre-training data such as ImageNet), can achieve comparable results in centralized settings and IID federated settings on most datasets, and outperform all ImageNet pre-training baselines in non-IID federated settings. 
This shows the potential of our framework to train high-quality federated models in real-world medical applications, where the data distribution across hospitals is typically non-IID.

Note that for the COVID-FL dataset, we have included three pre-training baselines using a large centralized in-domain dataset, CXR14, as previously mentioned.
As shown in Fig.~\ref{8c} and Table~\ref{tab:COVID-FL}, in the non-IID split of COVID-FL, both Fed-BEiT and Fed-MAE outperform the CXR14 supervised pre-training baseline (ViT CXR14). However, the improvement in performance of our proposed method compared to the CXR14 self-supervised pre-training baselines (MAE CXR14 and BEiT CXR14)is not significant. 
This suggests that, if a large centralized in-domain medical dataset is available, pre-training on it and fine-tuning with downstream target task data may be a good alternative to our proposed method. 
Nonetheless, it is rare for such datasets to exist for various medical tasks due to privacy and ownership concerns.

\subsubsection{More label-efficient}

%----------fig9 start --------
\begin{figure}[t]
\begin{center}
     \includegraphics[width=0.9\linewidth]{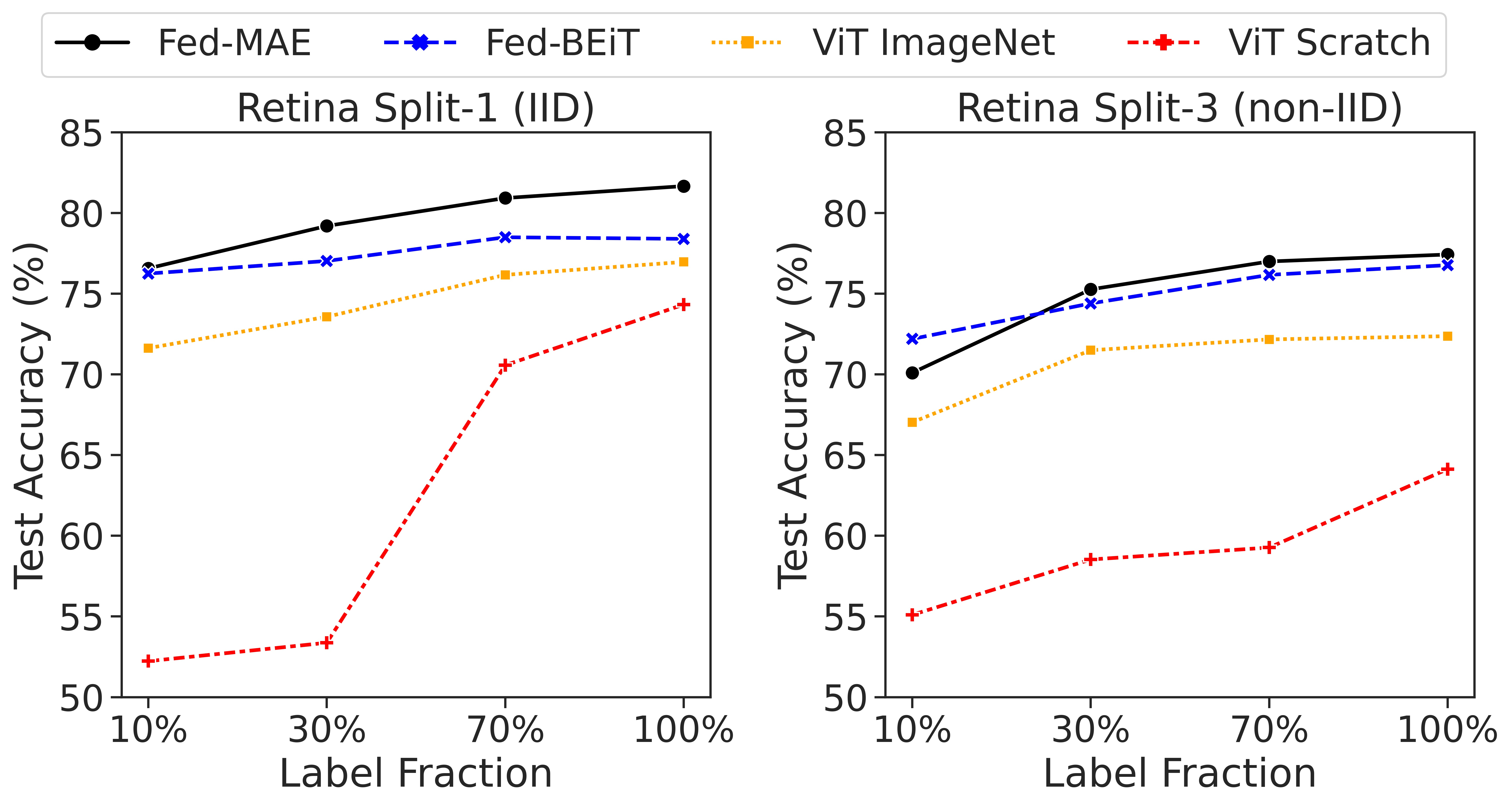}
\end{center}
\vspace{-1.0em}
\caption{Test accuracy for Retina under IID and non-IID settings with different fractions of labeled training samples.}
\label{Fig:scale}
\vspace{-.5em}
\end{figure}
%----------fig9 end --------

We conduct further experiments to evaluate the model performance under limited label scenarios using the Retina dataset.
Specifically, we reduce the number of labeled training images by different ratios during federated fine-tuning, taking approximately 70\%, 30\% and 10\% of the labeled samples from each class, resulting in a total of 6000, 3000 and 1000 labeled training data. 
Fig.~\ref{Fig:scale} shows the test accuracy when fine-tuning with different fractions of labeled data. 
In both IID and non-IID settings, our methods consistently improve the performance compared to the supervised baseline with ImageNet pre-training. 
The results also show that the model trained from scratch has unsatisfactory test accuracy when the number of labeled images is limited, such as less than $55\%$ when the number of labeled images is 1000.

% GENERALIZE TO UNSEEN CLIENTS?
\subsubsection{Generalization to out-of-distribution data} 
One desired property of a well-trained federated model is its ability to generalize to out-of-distribution data. We test the genralization of our proposed methods using Skin-FL and compare them to the baselines stated in~\cite{bdair2021fedperl} and our ViT supervised baselines. Fed-BEiT and Fed-MAE perform slightly better than the supervised baseline with ImageNet pre-training and notably better than all other methods (Table~\ref{tab:Skin-FL}).

%----------table5 start --------
\begin{table}[t]
\footnotesize
\centering
\caption{F1-score (\%) on Skin-FL dataset.}
\label{tab:Skin-FL}
\vspace{-0.5em}
\begin{tabular}{ccccc}
\toprule[0.15em]
    \multirow{2}{*}{Method} &
    \multirow{2}{*}{Backbone} &
    Pre-training &
    \multicolumn{2}{c}{Non-IID Split} \\
\cline{4-5}
 & &Dataset &\T Malignant &Benign \\
\midrule
FedAvg~\cite{yang2021federated} & ViT-B &\emph{None}  &15.4  &97.7 \\
FedAvg~\cite{yang2021federated} & EfficientNet &ImageNet &16.1 &97.4 \\
FedMatch~\cite{jeong2020federated} & EfficientNet &ImageNet &16.0 &97.3 \\
FedPerl~\cite{bdair2022semi} & EfficientNet &ImageNet &17.8  &97.4 \\
FedAvg~\cite{yang2021federated} & ViT-B &ImageNet &23.5  &98.1   \\
\midrule
Fed-BEiT & ViT-B &Skin-FL &\textbf{24.2} &\underline{98.4}  \\
Fed-MAE & ViT-B &Skin-FL &\underline{23.6} &\textbf{98.5} \\
\bottomrule[0.15em]
\end{tabular}
% \vspace{-.5em}
\end{table} 
%----------table5 end --------

\subsubsection{Comparison with prior FL methods}

In this section, we compare the performance of our proposed methods to previous FL methods in terms of (1) robustness to data heterogeneity and (2) label efficiency.

We first compare the performance of our methods to four federated contrastive learning baselines: FedEMA, FedBYOL, FedMoCo and FedMoCov3.
FedEMA~\cite{zhuang2021divergence} is the state-of-the-art federated self-supervised per-training method that has achieved the best performance on CIFAR10.
FedBYOL and FedMoCo are two baselines that combine BYOL~\cite{grill2020bootstrap} and MoCo~\cite{he2020momentum} with FedAvg~\cite{yang2021federated}.
ResNet-50 is used as the backbone for the above three baselines. 
We also implement a baseline based on MoCov3~\cite{chen2021empirical} with ViT-B as the backbone, referred to as FedMoCov3.
For the training details, the pre-training procedure lasts for 1000 communication rounds for FedBYOL and FedEMA, 300 and 150 rounds for FedMoCov3 and FedMoCo, respectively, until convergence. The AdamW optimizer is used with a batch size of 256 with a base learning rate of 0.03 for FedMoCo, and a batch size of 384 and a base learning rate starting at 5e-4 for the other three baselines.

As shown in Table~\ref{tab:fedssl_noniid}, while all of the self-supervised learning methods improve the robustness to data heterogeneity compared to random initialization, our proposed methods outperform all the baselines on the Retina dataset under the heterogeneous data partitions Split-2 and Split-3. 
Specifically, Fed-MAE and Fed-BEiT surpass the previous state-of-the-art FedEMA by 3.47\% and 2.81\%, respectively, in test accuracy for the severe non-IID Split-3. 
In addition, Table~\ref{tab:fedssl_label_ratio} compares the performance of our methods and the four baselines on the IID Split-1 of the Retina dataset with different fractions of labeled training samples. Our proposed methods outperform all the baselines with limited annotations, with greater improvements in performance when fine-tuning with fewer labeled samples.

%----------table6 start --------
\begin{table}[t]
\footnotesize
\centering
\caption{Test accuracy (\%) for federated fine-tuning on the Retina dataset using the proposed methods and federated self-supervised pre-training baselines.}
\vspace{-0.5em}
\label{tab:fedssl_noniid}
\begin{tabularx}{0.48\textwidth}{s e f f f f}
\toprule[0.15em]
Method
&Backbone
&Central
&Split1
&Split2
&Split3
\\
\midrule
\emph{Rand. init.}
&ViT-B
&73.70 &74.33 &69.50 &64.13 
\\
FedMoCov3
&ViT-B
&79.35 &78.06 &74.98 &72.32 
\\
FedMoCo
&ResNet-50
&77.50 &75.80 &73.03 &70.10
\\
FedBYOL
&ResNet-50
&80.10 &78.43 &75.27 &72.93
\\
FedEMA~\cite{zhuang2021divergence}
&ResNet-50
&\underline{80.12} &\underline{78.51} &76.08 &73.96\\
\midrule
Fed-BEiT
&ViT-B
&79.47 &78.40 &\underline{76.80} &\underline{76.77}
\\
Fed-MAE
&ViT-B
&\textbf{81.93} &\textbf{81.67} &\textbf{79.40} &\textbf{77.43}\\
\bottomrule[0.15em]
\end{tabularx}
\vspace{-.5em}
\end{table} 
%----------table6 end --------

%----------table7 start --------
\begin{table}[t]
\footnotesize
\centering
\caption{Test accuracy (\%) for federated fine-tuning on Retina Split-1 (IID) with different fractions of labeled samples.}
\vspace{-0.5em}
\label{tab:fedssl_label_ratio}
\begin{tabularx}{0.48\textwidth}{s e f f f f}
\toprule[0.15em]
Method
&Backbone
&10\%
&30\%
&70\%
&100\%
\\
\midrule
\emph{Rand. init.}
&ViT-B
&52.23
&53.37
&70.57
&74.33
\\
FedMoCov3
&ViT-B
&72.42 &73.34 &77.69 &78.06 
\\
FedMoCo
&ResNet-50
&65.20
&71.73
&75.20
&75.80
\\
FedBYOL
&ResNet-50
&72.77
&74.27
&77.10
&78.43
\\
FedEMA~\cite{zhuang2021divergence}
&ResNet-50
&72.95 &74.30 &77.02 &\underline{78.51} \\
\midrule
Fed-BEiT
&ViT-B
&\underline{76.25} &\underline{77.03} &\underline{78.50} &78.40
\\
Fed-MAE
&ViT-B
&\textbf{76.57} &\textbf{79.20} &\textbf{80.93} &\textbf{81.67}
\\
\bottomrule[0.15em]
\end{tabularx}
\end{table} 
%----------table7 end --------

%----------fig10 start (compare with prior FL methods)--------
\begin{figure}[t]
\begin{center}
% \vspace{-1em}
\includegraphics[width=0.9\linewidth]{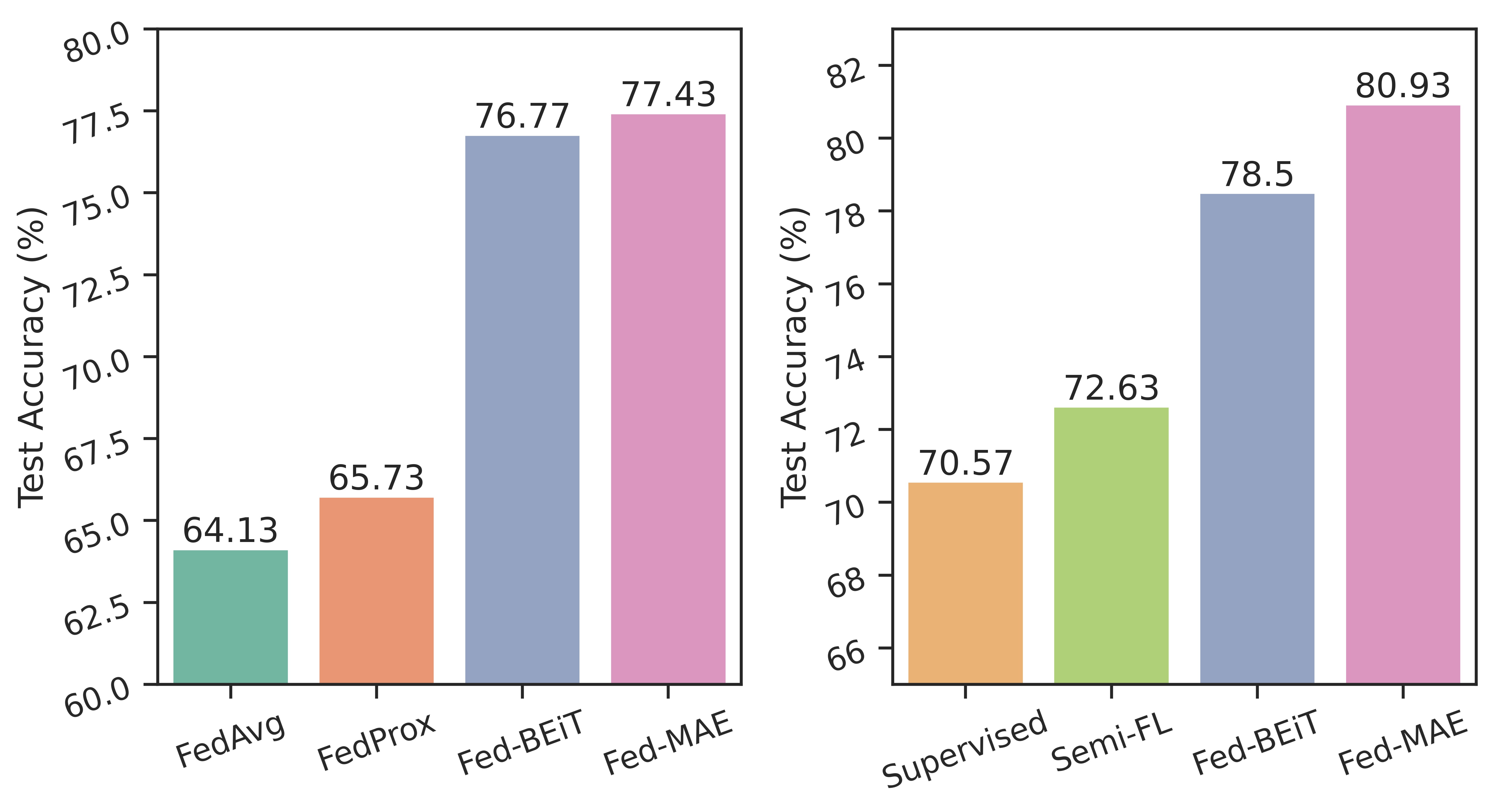} 
    \vskip -1.5em
    \subfloat[\label{10a}under non-IID settings]{\hspace{.45\linewidth}}
    \subfloat[\label{10b}under label-deficient settings]{\hspace{.45\linewidth}}
\end{center}
\vspace{-1.em}
\caption{Comparison of model performance under non-IID and label-deficient settings.} 
\label{Fig:compare}
% \vspace{-0.5em}
\end{figure}
%----------fig10 end --------

Additionally, we compare our methods to (1) FedProx~\cite{li2020federated} to evaluate their robustness to non-IID data, and to (2) semi-supervised FL (Semi-FL~\cite{yang2021federated}) to examine their effectiveness with limited labels. 
ViT-B is used as the backbone for FedProx and Semi-FL.

Fig.~\ref{10a} compares the test accuracy under the Retina dataset for the severe non-IID partition Split-3. 
To mitigate the weight divergence caused by data heterogeneity, FedProx~\cite{li2020federated} adds an $L_2$ regularization term $\frac{\mu}{2}||w-w_t||^2$ to the local objective function (Eq.~\ref{eq:local_objective}) during local client updates. 
We observe that training using FedProx can improve $1.60\%$ of accuracy from the FedAvg baseline after carefully tuning the optimization parameters $\mu$ ($\mu$ is set to 0.001). 
However, the gain from using our methods is significantly larger than using FedProx. 
Specifically, Fed-MAE yields a gain of $13.3\%$ in test accuracy. 
It is worth noting that the application of self-supervised pre-training in our method is orthogonal to optimization-based FL algorithms such as FedProx. 
Combining both could potentially further boost the model performance. 
% We leave this part as future work.
Fig.~\ref{10b} shows the test accuracy of different methods when training with 70\% labeled data on the IID Split-1 of the Retina dataset. 
To improve model performance when labeled data is scarce, Semi-FL~\cite{yang2021federated} leverages the unlabeled data in conjunction with supervision from the labeled data. 
It trains the clients with labeled data in a fully supervised manner for 400 epochs and then jointly trains with an extra client holding all the unlabeled data for another 400 epochs. 
This method was designed for segmentation tasks, and we adapt it to classification tasks. 
For the unlabeled client, we use a consistency loss function based on data augmentation, calculating the cross-entropy loss between the outputs of the augmented data and the pseudo labels based on the predictions of the original data. 
According to Fig.~\ref{10b}, our Fed-MAE outperforms the supervised baseline by $10.36\%$ and the semi-supervised method by $8.3\%$.

%+++++++++++++++++++++++ 4-4: ABLATION STUDIES
\subsection{Ablation Studies}
\label{sec:ablation}
We perform ablation studies to assess the impact of factors such as communication rounds, training data size, mask ratios, and data augmentations on model performance.

\subsubsection{Number of communication rounds}
\begin{figure}[t]
\begin{center}
\includegraphics[width=0.95\linewidth]{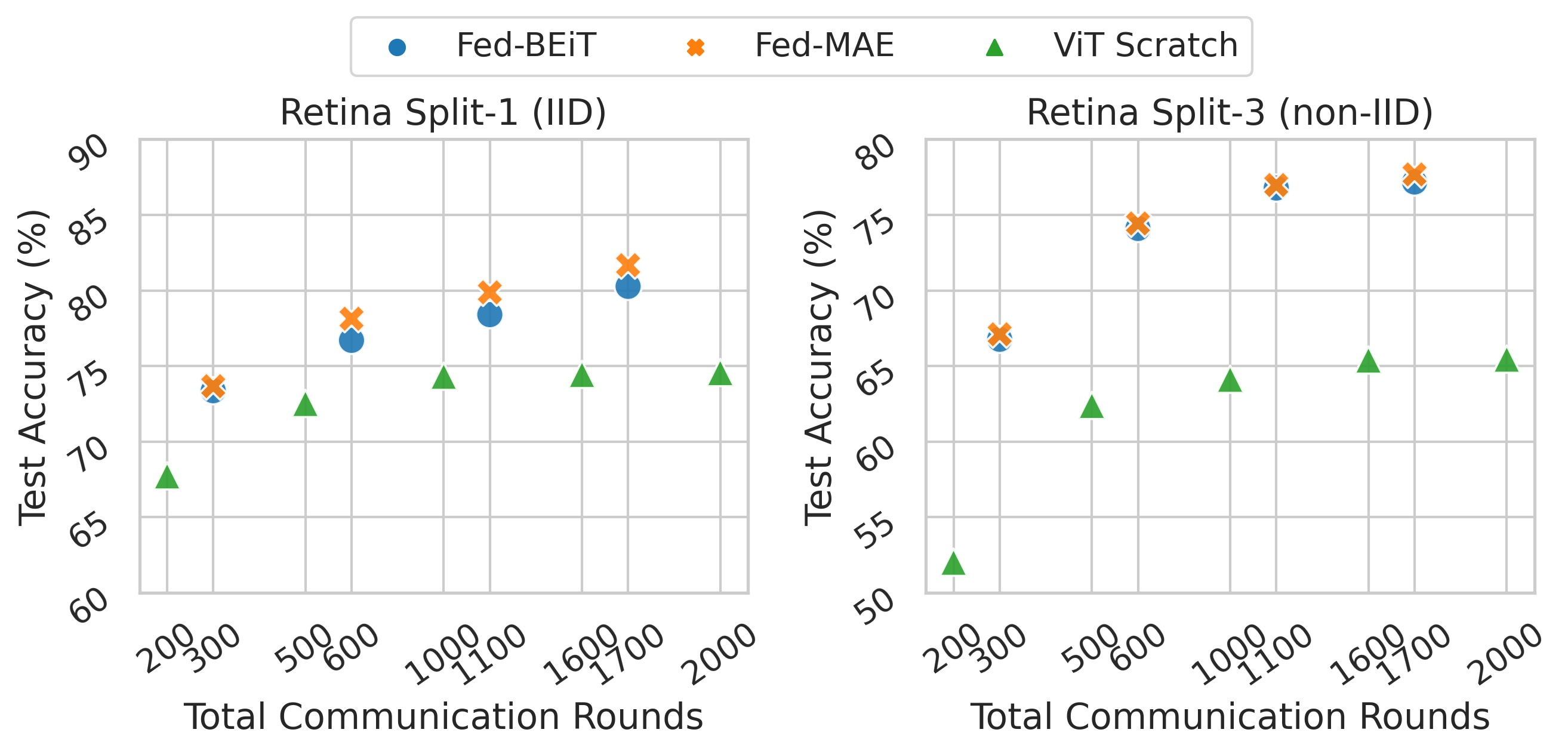}
\end{center}
\vspace{-1em}
\caption{Ablation study on the number of total communication rounds using Retina Split-1 (IID) and Split-3 (severe non-IID) partitions.}
\label{Fig:cost}
\vspace{-.5em}
\end{figure}

Fig.~\ref{Fig:cost} compares the accuracy of our methods (Fed-BEiT and Fed-MAE) and the baseline (ViT Scratch) for different numbers of total communication rounds ($T_\text{total} = T_p + T_f$), where $T_p$ and  $T_f$ represent the number of communications rounds for pre-training and fine-tuning, respectively.
For the proposed methods, we use $T_p\in\{200, 500, 1000, 1600\}$ for pre-training and $T_f=100$ for fine-tuning, while for the baseline without pre-training, we use $T_p=0$ and $T_f\in\{200, 500, 1000, 1600, 2000\}$.

%----------fig11 start --------
%----------fig11 end --------

On the IID (Split-1) and severe non-IID (Split-3) partitions of the Retina dataset, our methods consistently outperform the baseline. 
For the baseline without pre-training (represented by the green markers in Fig.~\ref{Fig:cost}), the accuracy remains below 75\% for Split-1 and 65\% for Split-3, even when $T_f$ reaches 2000.
By using our proposed pre-training techniques (represented by the blue and orange markers in Fig.~\ref{Fig:cost}), the accuracy for the two splits surpasses 75\% and 65\% with much shorter $T_\text{overall}$ (less than 600 for Split-1 and less than 300 for Split-3), and continues to increase as $T_p$ increases (reaching around 81\% and 77\% for Split-1 and Split-3, respectively, when $T_p$ increases to 1600).

Here, we examine the trade-off between model accuracy and communication cost (the number of communication
rounds $\times$ the size of the communicated model). 
During fine-tuning, the size of the communicated model (encoder + linear classifier, 85.8M parameters) is consistent for both our methods and the baseline, while the model communicated during pre-training (encoder + decoder) is larger, with 111.7M parameters for Fed-MAE and 92M parameters for Fed-BEiT, which are 1.3$\times$ and 1.11$\times$ the size of the model transmitted during fine-tuning.
With the size of the communicated model
taken into consideration, our proposed methods still outperform the baseline.
For example, when $T_p$ is 500 and $T_f$ is 100, our method Fed-BEiT achieves an accuracy of 74.10\% on Retina Split-3 with a communication cost of 500$\times$92M+100$\times$85.8M, while the baseline achieves an accuracy of about 63\% with the same communication cost at a communication round of 636.

%----------table8 start --------
\begin{table}[t]
\footnotesize
\centering
\vspace{-.2em}
\caption{Ablation study on the size of the training data.}
\vspace{-.7em}
\label{tab:training_data_size}
\begin{tabular}{cccc}
\toprule[0.15em]
Number of training images
&3000
&6000
&9000\\
\hline
\T 
Fed-BEiT
&69.93
&74.03
&76.77\B
\\
Fed-MAE
&71.70
&75.67
&77.43
\\
\hline
\T
Init. w/ ImageNet weights 
&71.50
&72.17
&72.37
\\
\bottomrule[0.15em]
\end{tabular}
% \vspace{-.5em}
\end{table} 
%----------table8 end --------

\subsubsection{Training data size}
To investigate the effect of the size of the training data on model performance, we reduce the number of total training samples of the Retina Split-3 dataset from 9000 to 6000 and 3000. Table~\ref{tab:training_data_size} compares the accuracy of our methods and ImageNet pre-training.
We find the performance gap between the model pre-trained using our methods and the model initialized with ImageNet weights is reduced when the number of total training images decreases. 
Therefore, it is applicable to directly use pre-trained ImageNet weights when the total number of training images from all clients combined is limited (\emph{e.g.}, less than 3000).

%----------table9 start --------
\begin{table}[t]
\footnotesize
\centering
\vspace{-.2em}
\caption{Ablation study on mask ratio.}
\label{tab:mask_ratio}
\vspace{-.7em}
\begin{tabular}{cc|ccccc}
    \toprule[0.15em]
    \multicolumn{1}{c}{}
    &\multicolumn{1}{c|}{}
    &\multicolumn{5}{c}{\centering\arraybackslash Mask Ratio} \B\\
  \hline
  \T
  Dataset
    &\multicolumn{1}{c|}{Method}
    &\multicolumn{1}{c}{30\%} 
    &\multicolumn{1}{c}{40\%} &\multicolumn{1}{c}{50\%}
    &\multicolumn{1}{c}{60\%} &\multicolumn{1}{c}{70\%} \B\\
 \hline
 \T
 \multirow{2}{*}{Retina} 
    &\multicolumn{1}{c|}{BEiT}
    &\multicolumn{1}{c}{78.53} &\multicolumn{1}{c}{\textbf{79.47}} &\multicolumn{1}{c}{79.00}
    &\multicolumn{1}{c}{78.37} &\multicolumn{1}{c}{77.60} \\
    &\multicolumn{1}{c|}{MAE}
    &\multicolumn{1}{c}{79.73} &\multicolumn{1}{c}{81.50} &\multicolumn{1}{c}{81.73}
    &\multicolumn{1}{c}{\textbf{81.93}} &\multicolumn{1}{c}{80.90} \B\\\hline
  \T
  \multirow{2}{*}{Derm} 
    &\multicolumn{1}{c|}{BEiT}
    &\multicolumn{1}{c}{92.62} &\multicolumn{1}{c}{\textbf{93.27}} &\multicolumn{1}{c}{92.57}
    &\multicolumn{1}{c}{92.04} &\multicolumn{1}{c}{91.83} \\
    &\multicolumn{1}{c|}{MAE}
    &\multicolumn{1}{c}{92.94} &\multicolumn{1}{c}{93.35} &\multicolumn{1}{c}{93.60 }
    &\multicolumn{1}{c}{\textbf{93.79}} &\multicolumn{1}{c}{93.07} \B\\\hline
  \T
  \multirow{2}{*}{COVID-FL} 
    &\multicolumn{1}{c|}{BEiT}
    &\multicolumn{1}{c}{95.79} &\multicolumn{1}{c}{\textbf{95.84}} &\multicolumn{1}{c}{95.67}
    &\multicolumn{1}{c}{95.57} &\multicolumn{1}{c}{95.32} \\
    &\multicolumn{1}{c|}{MAE}
    &\multicolumn{1}{c}{\textbf{96.55}} &\multicolumn{1}{c}{96.53} &\multicolumn{1}{c}{96.23}
    &\multicolumn{1}{c}{95.80} &\multicolumn{1}{c}{95.67} \\
\bottomrule[0.15em]
\end{tabular}
\vspace{-.5em}
\end{table}
%----------table9 end --------

\subsubsection{Masking ratio}
We examine the optimal mask ratio for medical datasets: Retina, Derm and COVID-FL (Table~\ref{tab:mask_ratio}).
Our results indicate that the optimal mask ratio for BEiT and MAE on Retina and Derm is 40\% and 60\%, respectively, which is consistent with previous findings on ImageNet~\cite{bao2021beit,he2021masked}. 
We also observe that the optimal mask ratio for MAE on COVID-FL is 30\%. This makes sense because retina and skin images in the Retina and Derm datasets are more similar to natural images than the chest X-ray images in the COVID-FL dataset.

%----------table10 start --------
\begin{table}[t]
\footnotesize
\centering
\vspace{-0.2em}
\caption{Ablation study on data augmentation.}
\vspace{-0.5em}
\label{tab:data_aug}
\begin{tabularx}{0.46\textwidth}{c|c}
\toprule[0.15em]
Data Augmentation & Test Accuracy \B\\
\hline
\T random crop + horiz. flip & 81.03  \\
random crop + horiz. flip + gray scale + color jitter& \textbf{81.73}  \\
\bottomrule[0.15em]
\end{tabularx}
% \vspace{-1.3em}
\end{table}
%----------table10 end --------

\subsubsection{Data Augmentation}
Table~\ref{tab:data_aug} shows the effect of data augmentation on MAE pre-training in centralized settings using the Retina dataset.
Adding gray scaling and color jittering improves accuracy by 0.7\%, suggesting that task-specific data augmentations may be beneficial for pre-training on medical tasks.

\section{Conclusion}
In this paper, we propose a privacy-preserving and federated self-supervised learning framework that collaboratively trains models on decentralized data using masked image modeling as the self-supervised task. 
Our framework is robust to non-IID data distribution across clients, and significantly outperforms state-of-the-art ImageNet supervised pre-training baselines under severe data heterogeneity.
It also generalizes well to out-of-distribution data and effectively learns with limited labeled data.
Across diverse medical datasets, we show that our proposed method outperforms existing federated self-supervised learning methods, as well as optimization-based and semi-supervised FL methods, under non-IID and label deficient scenarios.

% \clearpage
\bibliographystyle{IEEEtran}
\bibliography{tmi}
\iffalse

\fi

\end{document}